\documentclass[twoside]{article}

\usepackage[accepted]{aistats2023}
\usepackage{amsmath}
\usepackage{color} 
\usepackage{mathabx}
\usepackage{stfloats}
\usepackage{float}
\usepackage{lipsum} 
\usepackage{fancyhdr}
\usepackage{algorithmic}
\usepackage{textcomp}
\usepackage{xcolor}
\usepackage{tikz}
\usepackage{calc}
 
\usepackage{etoolbox}
\usepackage{algorithm} 
\usepackage{algorithmic}
\usepackage{wrapfig}

\usepackage{graphicx}
\usepackage{float}
\usepackage{booktabs}
\usepackage{multirow}
\usepackage{multicol}
\usepackage{hyperref}
\usepackage{subcaption}

\usepackage{comment}
\usepackage{ifthen}

\newcommand{\boldhdr}[1]{\noindent \textbf{#1.}}
\newcommand{\showcomments}{yes}
\newcommand\m[1]{
    \ifthenelse{\equal{\showcomments}{yes}}{{\color{magenta} [Ming: #1]}}{\ignorespaces}
}

\newcommand{\kaiqi}[1]{\textcolor{black}{#1}}

\begin{document}

%

%

\twocolumn[

\aistatstitle{Automatic Attention Pruning: Improving and Automating Model Pruning using Attentions}

\aistatsauthor{ Kaiqi Zhao \And Animesh Jain \And  Ming Zhao }

\aistatsaddress{ Arizona State University \And  Meta \And Arizona State University } ]


\begin{abstract}
Pruning is a promising approach to compress deep learning models in order to deploy them on resource-constrained edge devices.
However, many existing pruning solutions are based on unstructured pruning, which yields models that cannot efficiently run on commodity hardware; and they often require users to manually explore and tune the pruning process, which is time-consuming and often leads to sub-optimal results. 
To address these limitations, this paper presents Automatic Attention Pruning (AAP), an adaptive, attention-based, structured pruning approach to automatically generate small, accurate, and hardware-efficient models that meet user objectives. 
First, it proposes iterative structured pruning using activation-based attention maps to effectively identify and prune unimportant filters. Then, it proposes adaptive pruning policies for automatically meeting the pruning objectives of accuracy-critical, memory-constrained, and latency-sensitive tasks. A comprehensive evaluation shows that AAP substantially outperforms the state-of-the-art structured pruning works for a variety of model architectures. 
Our code is at: \url{https://github.com/kaiqi123/Automatic-Attention-Pruning.git}.
\end{abstract}

\section{Introduction}\label{sec:introduction}

Deep neural networks (DNNs) have substantial computational and memory requirements. As the use of deep learning grows rapidly on a wide variety of Internet of Things and devices, the mismatch between resource-hungry DNNs and resource-constrained devices also becomes increasingly severe.
Pruning is a promising approach to identify and remove the parameters that do not contribute significantly to the accuracy of a DNN.
Recent works based on the Lottery Ticket Hypothesis (LTH) have achieved great results in creating smaller and more accurate models (dubbed as ``winning tickets'') through iterative pruning with rewinding~\cite{frankle2018lottery}. 
However, LTH has only been shown to work with unstructured pruning which, unfortunately, leads to models with low sparsity and difficult to accelerate on commodity hardware; e.g., directly applying NVIDIA cuSPARSE on unstructured pruned models can lead to a 60$\times$ slowdown compared to dense kernels on GPUs~\cite{deftnn}.
Moreover, most pruning methods require users to explore and adjust multiple hyper-parameters, which is time-consuming and often leads to sub-optimal results; e.g., with LTH-based iterative pruning, users need to determine how many parameters to prune in each pruning round. 

We propose Automatic Attention Pruning (AAP), an \textit{adaptive}, \textit{attention-based}, \textit{structured pruning} solution to automatically generate small, accurate, and hardware-efficient models that meet users' accuracy, size, and speed requirements.
We improve the LTH-based iterative pruning framework by proposing two methods.
First, we propose a novel attention pruning method to identify and remove unimportant filters. Specifically, we properly define an attention mapping function that takes the 2D activation feature map of a filter as input and outputs a 1D value used to indicate the importance of the filter. This approach is more effective than weight-value-based filter pruning~\cite{renda2020comparing, zhuang2020neuron, wang2019eigendamage} because activation-based attention values not only capture the features of inputs but also contain the information of convolution layers that act as feature detectors for prediction tasks. Also, it is better than previous activation-based filter pruning methods~\cite{lin2020hrank, liu2017learning} since the accuracy of its measurement does not depend on the amount of inputs. 

Second, we propose an adaptive pruning method that automatically optimizes the pruning process according to different user objectives. For latency-sensitive scenarios like interactive virtual assistants, we propose FLOPs-guaranteed pruning to achieve the best accuracy with the acceptable inference speed; for memory-limited environments like embedded systems, we propose model-size-guaranteed pruning to achieve the best accuracy and fit the memory constraint; for accuracy-critical applications such as those on self-driving cars, we propose accuracy-guaranteed pruning to create the most resource-efficient model with the acceptable accuracy loss. Given the target, our method adaptively controls the pruning aggressiveness by adjusting the global threshold used to prune filters.
Moreover, it recognizes the difference in each layer's contribution to the model's size and computational complexity and uses a layer-wise threshold, calculated by dividing each layer's remaining parameters or FLOPs by the entire model's remaining parameters or FLOPs, to prune each layer with a differentiated level of aggressiveness.

The proposed AAP outperforms the related works significantly in all cases targeting accuracy loss, parameters reduction, and FLOPs reduction for a variety of model architectures. 
For example, on ResNet-56 with CIFAR-10, without accuracy drop, AAP achieves the largest parameters reduction (79.11\%), outperforming the related works by 22.81\% to 66.07\%, and the largest FLOPs reduction (70.13\%), outperforming the related works by 14.13\% to 26.53\%. 
On ResNet-50 with ImageNet, for the same level of parameters and FLOPs reduction, AAP achieves the smallest accuracy loss, lower than the related works by 0.08\% to 2.61\%; 
and for the same level of accuracy loss, AAP reduces significantly more parameters (6.45\% to 29.61\% higher than the related works) and more FLOPs (0.82\% to 17.2\% higher than the related works).

In summary, our main contributions are: 1) a novel iterative, structured pruning approach for finding the ``winning ticket'' models that are hardware efficient; 2) a new attention-based mechanism for accurately identifying unimportant filters for pruning, which is much more effective than existing methods; and 3) an adaptive pruning method that can automatically optimize the pruning process according to diverse real-world scenarios.
\section{Background and Related Works}
\label{sec:background}

\noindent{\textbf{Unstructured Pruning vs. Structured Pruning.}} 
Unstructured pruning (e.g.,~\cite{lecun1990optimal,han2015deep,molchanov2017variational}) prunes individual elements in the weight tensors of a model. It has less impact on model accuracy, compared to structured pruning, because it is finer-grained, but unstructured pruned models are hard to accelerate on commodity hardware.
Structured pruning is a coarser-grained approach that prunes entire regular regions of the weight tensors of a model. 
It is more difficult to prune a model without causing accuracy loss using structured pruning, because by removing entire regions, it might remove weight elements that are important to the final accuracy. However, structured pruned models can be mapped easily to general-purpose hardware and accelerated directly with off-the-shelf hardware and libraries~\cite{he2018amc}.

\smallskip
\boldhdr{One Shot Pruning vs. Iterative Pruning} 
One-shot pruning prunes a pre-trained model and then retrains it once, whereas iterative pruning prunes and retrains the model in multiple rounds. 
Iterative pruning generally achieves much better performance than one-shot pruning because multiple retraining phases help recover the accuracy lost during pruning.
In particular, recent works based on the Lottery Ticket Hypothesis (LTH) have achieved great results in creating smaller and more accurate models through iterative pruning with rewinding~\cite{frankle2018lottery}. LTH posits that a dense network has a sub-network, termed as a ``winning ticket'', which can achieve an accuracy comparable to the original network. 
However, existing LTH-based works consider only unstructured pruning, e.g., Iterative Magnitude Pruning (IMP)~\cite{frankle2018lottery, frankle2019stabilizing} and \kaiqi{Synflow~\cite{tanaka2020pruning}}, which, as discussed above, are hardware-inefficient. 

\smallskip
\boldhdr{Automatic Pruning} 
For pruning to be useful in practice, it is important to automatically meet the pruning objectives for diverse machine learning applications and devices. Most pruning methods require users to explore and tune multiple hyper-parameters, e.g., with LTH-based pruning, users need to determine how many parameters to prune in each round. Manual tuning is time-consuming and often leads to sub-optimal results. 
Some works use learning-based methods to find smaller models in a Neural Architecture Search (NAS) approach: AMC \cite{he2018amc}, NAS \cite{zoph2018learning}, NT \cite{caireinforcement}, and N2N \cite{ashok2017n2n} use reinforcement learning, and GAL \cite{lin2019towards} uses adversarial learning.
But these methods have to explore a large search space of all available layer-wise sparsity, which is time consuming when neural networks are large and datasets are complex.

Therefore, there is a great need for an automatic, iterative, structured pruning solution that can automatically and efficiently generate small, accurate, and hardware-efficient models.
The challenges are three-fold.

First, \textit{how to effectively identify the insignificant parameters in a model to prune?} Existing works have explored different mechanisms, e.g., L2-Norm in \kaiqi{Soft Filter Pruning (SFP) \cite{he2018soft}, Soft Channel Pruning (SCP) \cite{kang2020operation} and EagleEye \cite{li2020eagleeye}}, geometric median in FPGM \cite{he2019filter}, Hessian in \kaiqi{EigenDamage \cite{wang2019eigendamage}}, Empirical Sensitivity in \kaiqi{Provable Filter Pruning (PFP)} \cite{liebenwein2019provable}, adversarial knockoff features in SCOP \cite{tang2020scop}, polarization regularizer in \kaiqi{Neuron-level Structured Pruning (NSP) \cite{zhuang2020neuron}}, LASSO regression in \kaiqi{Channel Pruning (CP) \cite{he2017channel}}, and other information considering the relationship between neighboring layers (\kaiqi{Gate Batch Normalization (GBN) \cite{you2019gate}}, Sparse Structure Selection (SSS) \cite{huang2018data}, Hinge \cite{li2020group}, \kaiqi{Pruning From Scratch (PFS) \cite{wang2020pruning} and Stripe-Wise Pruning (SWP) \cite{meng2020pruning}}). 
In comparison, \textit{activation-based attention}, proposed in this paper, can more effectively capture the importance of filters, and pruning based on attention values can produce much better models, as quantitatively shown in our evaluation (Section~\ref{sec:evaluation}).

Second, \textit{how to design an effective iterative pruning process to recover the accuracy loss caused by structured pruning?} 
LTH-based iterative pruning is a promising approach, but it has only been shown to work with unstructured pruning such as IMP. Its counterpart in structured pruning---Iterative L1-norm-based pruning (ILP)~\cite{renda2020comparing}, which removes filters based on their L1-norm values, cannot effectively prune a model while maintaining its accuracy. For example, ILP can prune ResNet-50 by at most 11.5\% of parameters when the maximum accuracy loss is limited to 1\% on ImageNet. So directly applying iterative pruning with existing weight-magnitude-based structured pruning methods does not produce accurate pruned models. 
This paper proposes a novel \textit{LTH-based iterative, structured pruning} solution using attentions, and it significantly outperforms ILP and other related structured pruning works that involve an iterative process (GDP~\cite{guo2021gdp}, \kaiqi{ACTD~\cite{wang2021accelerate}, Quantization and Pruning (QP)~\cite{paupamah2020quantisation}, IMP-Refill and IMP-Regroup~\cite{chen2022coarsening})}.

The third challenge is \textit{how to automate the pruning process so it does not require any human intervention?} 
The existing structured pruning works all require difficult hand-tuning of many hyper-parameters,
e.g., DCP \cite{zhuang2018discrimination} and MDP \cite{guo2020multi} need multiple hyper-parameters to balance the original task-specific loss and the additional pruning loss; VCNNP \cite{zhao2019variational} requires careful settings of $\tau$ and $\theta$ to decide which filters to prune; DMC \cite{gao2020discrete} and \kaiqi{DeepHoyer \cite{yang2019learning}} require parameters to decide the regularization strength with different settings for different datasets and models.
To address this challenge, this paper proposes a fully automated pruning solution that can automatically generate pruned models that meet users' diverse model accuracy, size, and speed requirements.

\section{Methodology}\label{sec:methodology}

\begin{figure}[t]
    \begin{algorithm}[H]
        \small
        \caption{Adaptive Iterative Structured Pruning}
        \begin{algorithmic}[1]
            \STATE \textbf{Input:} An uncompressed network, and the pruning target
            \STATE \textbf{Output:} A pruned network that meets the target
            \STATE {[}Initialize{]} Initialize a network $f(x; M^0 \odot W_{0}^{0})$ with the initial mask $M^0=\left \{ 0,1 \right \}^{\left | W_{0}^{0} \right |}$ 
            \STATE {[}Save weights{]} Train the network for $k$ epochs, yielding network $f(x; M^0 \odot W_{k}^{0})$, and save weights $W_{k}^{0}$
            \STATE {[}Train to converge{]} Train the network for \kaiqi{$E-k$} epochs to converge,  producing network \kaiqi{$f(x; M^0 \odot W_{E}^{0})$}
            \FOR {pruning round $r$ ($r \geq 1$)}
                \STATE {[}Calculate attention{]} Calculate the attention value of each filter using the attention mapping function $F(\cdot)$
                \STATE {[}Prune{]} From \kaiqi{$W_{E}^{r-1}$}, prune filters  with an attention value less than $T[r]$, producing a mask $M^r$ and a network \kaiqi{$f(x; M^r \odot W_{E}^{r-1})$}
                \STATE {[}Rewind Weights{]} Reset the remaining filters to $W_{k}^{0}$ at epoch $k$, producing network $f(x; M^r \odot W_{k}^{r-1})$
                \STATE {[}Rewind Learning Rate{]} Reset the learning rate schedule to its state from epoch $k$
                \STATE {[}Retrain{]} Retrain the unpruned filters for \kaiqi{$E-k$} epoch to converge, yielding network \kaiqi{$f(x; M^r \odot W_{E}^{r})$}
                \STATE {[}Evaluate{]} Evaluate the retrained network $f(x; M^r \odot W_{E}^{r})$ according to the target
                \STATE {[}Reset Weights{]} If the target is not met, reset the weights to an earlier round
                \STATE {[}Adapt Threshold{]} Calculate the next threshold $T[r+1]$ 
            \ENDFOR
        \end{algorithmic}
        \label{alg:overall_pruning}
    \end{algorithm}
\end{figure}

\begin{figure}[t]
	\begin{subfigure}{0.9\columnwidth}
		\centering
		\includegraphics[width=6cm]{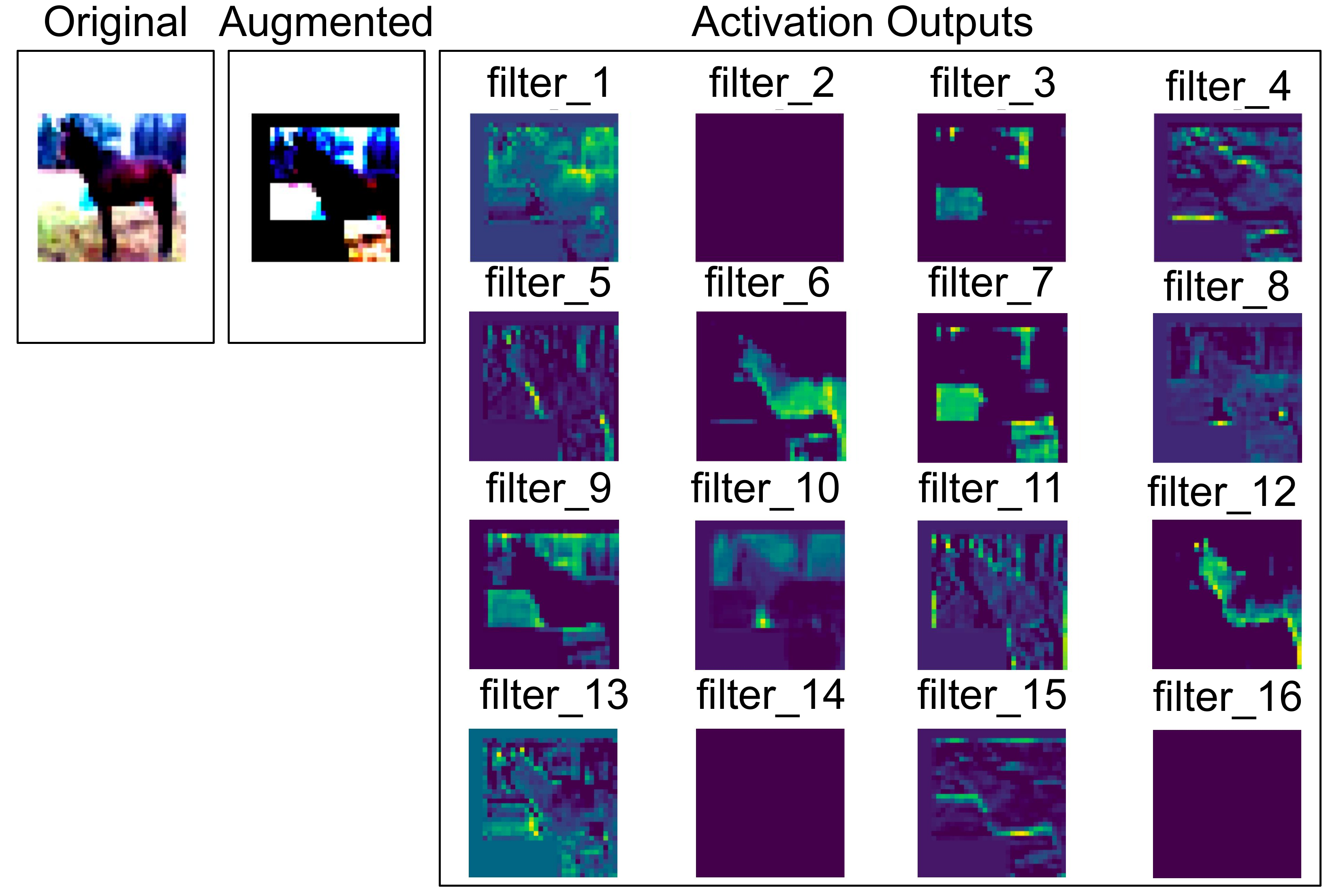}
		\caption{Filter activation outputs.}
		\label{fig:visualization_of_activations}
	\end{subfigure}
	\begin{subfigure}{0.9\columnwidth}
		\centering
        \includegraphics[width=6cm]{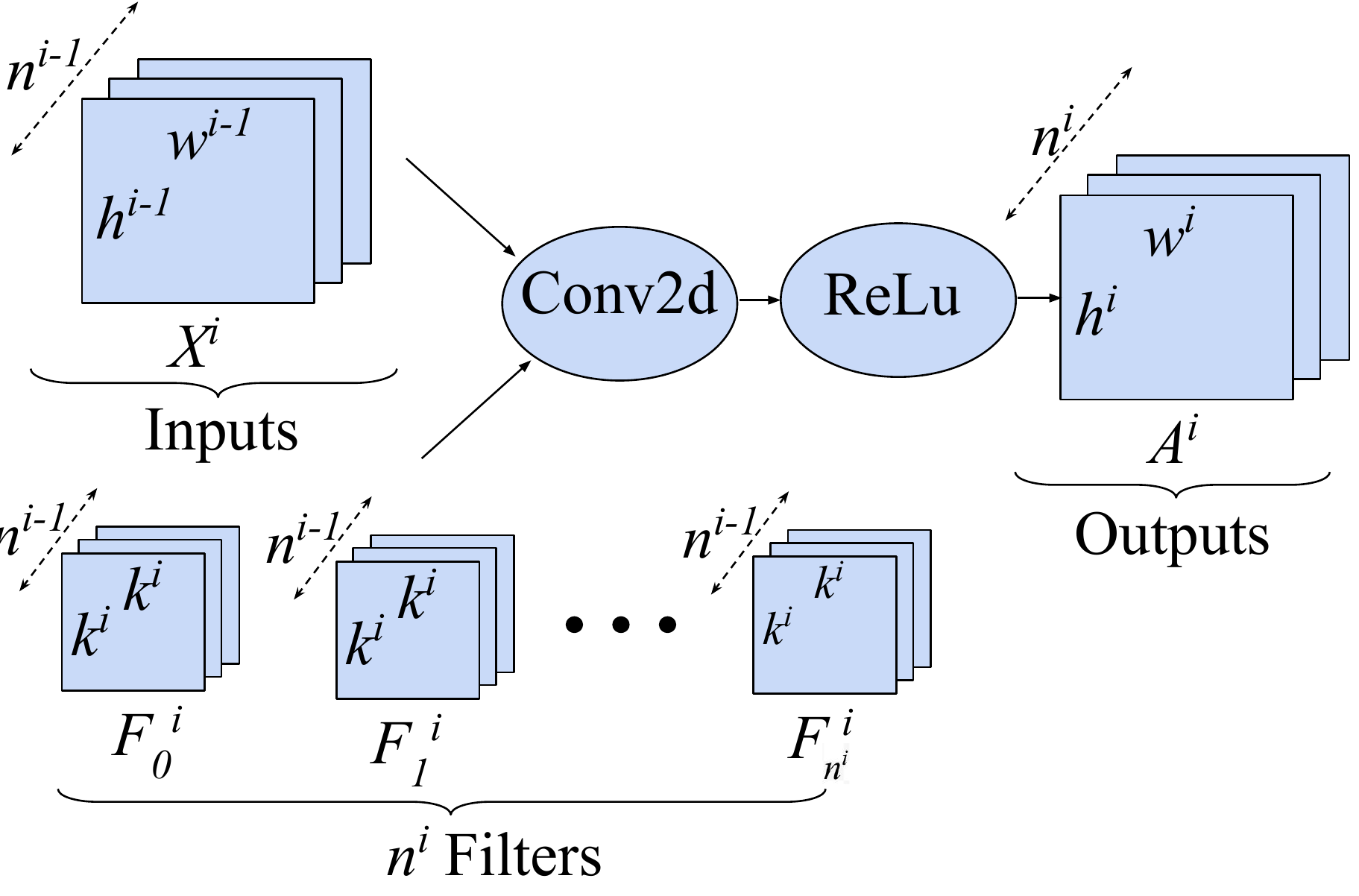}
		\caption{Convolution layer tensors.}
		\label{fig:conv2d}
	\end{subfigure}
	\caption{\kaiqi{(a) Activation outputs of 16 filters, and (b) input and output tensors of a convolution layer}.}
\end{figure}

Algorithm~\ref{alg:overall_pruning} lists the proposed adaptive structured pruning for AAP. We improve LTH-based iterative pruning by proposing activation-based attention pruning (Lines 7 and 8) and adaptive pruning policies (Lines 11--14), to automatically and efficiently generate a pruned model that meets the user's different objectives. 
To represent pruning of weights, \kaiqi{we use a mask $M^r \epsilon \left \{0, 1\right \} ^ {|W^r|}$ for each weight tensor $W_t^r$, where $r$ is the pruning round number and $t$ is the training epoch}. Therefore, the pruned network at the end of training epoch \kaiqi{$E$} is represented by the element-wise product \kaiqi{$M^r \odot W_E^r$}. Lines 3--5 are to train the original model to completion while saving the weights at epoch $k$. Lines 6--15 represent a pruning round. Lines 7 and 8 prune the model (discussed in Section~\ref{sec:filterPruning}). Lines 9 (optional) and 10 perform rewinding. Line 11 retrains the pruned model for the remaining \kaiqi{$E-k$} epochs. Line 12 evaluates the pruned model according to the pruning target. If the target is not met, Line 13 resets the weights to an earlier round. Line 14 calculates the threshold for the next pruning round following the adaptive pruning policy (discussed in Section~\ref{sec:adaptive_pruning})

\subsection{Attention-based Filter Pruning}
\label{sec:filterPruning}

First, we propose that, compared with the \textit{weight values} of a filter, its \textit{activation values} are more effective indicators of finding unimportant filters to prune.
Activations like ReLu enable non-linear operations, and enable convolutional layers to act as feature detectors.
If an activation value is small, then its corresponding feature detector is not important for prediction tasks. 
On the other hand, some filters, even though their weight values are small, can still produce useful non-zero activation values that are important for learning features during backpropagation.
We present a visual motivation in Figure~\ref{fig:visualization_of_activations}. 
The figure shows the activation outputs of 16 filters of a convolution layer on one input image. 
The first image on the left is the original image, and the second image is the input features after data augmentation. 
We observe that some filters extract image features with high activation patterns, e.g., the $6$th and $12$th filters. 
In comparison, the activation outputs of some filters are close to zero, such as the $2$nd, $14$th, and $16$th filters. 
Therefore, from visual inspection, removing filters with weak activation patterns is likely to have a low impact on the final accuracy of the pruned model.


Thus the key problem is to design a proper function that can reflect the useful information of the activation feature maps of each filter. 
%
%
\kaiqi{Previous activation-based filter pruning methods address this problem using different forms of activations: NN Slimming \cite{liu2017learning} measures Average Percentage of Zeros (APoZ) of the activations; HRank \cite{lin2020hrank} conductes a Singular Value Decomposition (SVD) for activations; AP+Coreset \cite{dubey2018coreset} computes the mean value of activations; Provable Filter Pruning (PFP) \cite{liebenwein2019provable} computes the sensitivity of activations. 
However, these data-driven approaches require a large amount of inputs (e.g., 50,000 of ImageNet images for NN Slimming and all training samples for AP+Coreset) to achieve a reasonably accurate prediction and leads to data-dependent compressed models.
Instead, we aim to design a function that is data-independent and robust to inputs: no matter what the input image is, the filters that can extract useful image features should always be maintained.}

Activation-based attention is a good indicator of neurons regarding their ability to capture features \cite{zagoruyko2016paying}. 
Attention has been proven to be useful in various tasks, including neural machine translation \cite{bahdanau2014neural}, object localization \cite{oquab2015object}, and knowledge transfer for image classification~\cite{zagoruyko2016paying}. 
\kaiqi{Also, motivated by human attention mechanism theories, attention maps can be obtained by computing a Jacobian of network outputs regarding the inputs \cite{simonyan2013deep}, guided backpropagation~\cite{springenberg2014striving}, or converting the linear classification layer into a convolutional layer \cite{zhou2016learning}.}
%
However, the effectiveness of using attention as a mechanism for model compression is currently unexplored, and it imposes new challenges, e.g., the attention mapping function defined in Attention Transfer~\cite{zagoruyko2016paying} outputs a flattened 2D matrix representing the ability to capture features of the whole convolution layer, not individual filters in that layer, and cannot be used to prune individual filters.

We address the challenge of effectively identifying insignificant filters in a network with novel designs for the attention mapping function.
We start with some notations shown in Figure~\ref{fig:conv2d}.
For the $i$th 2D convolution (conv2d) layer, let $X^i \epsilon R^{n^{i-1}\times h^{i-1}\times w^{i-1}}$ denote the input features, and $F^i_j  \epsilon R^{n^{i-1}\times k^i \times k^i}$ be the $j$th filter, where $h^{i-1}$ and $w^{i-1}$ are the height and width of the input features, respectively, $n^{i-1}$ is the number of input channels, $n^{i}$ is the number of output channels, and $k^i$ is the kernel size of the filter. The activation of the $j$th filter $F^i_j$ after ReLu mapping is therefore denoted by $A^i_j \epsilon R^{h^i \times w^i}$. 
%
The proposed attention mapping function takes a 2D activation $A^i_j \epsilon R^{h^i \times w^i}$ of filter $F^i_j$ as input, and outputs a 1D value which will be used as an indicator of the importance of filters. We consider three forms of activation-based attention mapping functions, where $p \geq 1$ and $a_{k,l}^i$ denotes every element of $A^i_j$:
\begin{enumerate} 
  \item \textbf{Attention Mean} (mean of the \kaiqi{activation} values) 
  
  $F_{mean}(A^i_j)= \frac{1}{h^{i}\times w^{i}}\sum_{k=1}^{h^i}\sum_{l=1}^{w^i} {\left | a_{k,l}^i  \right |}^{p}$; 
  \item \textbf{Attention Max} (max of the \kaiqi{activation} values) 
  
  $F_{max}(A^i_j)= max_{l=1, h^{i}\times w^{i}} \left | a_{k,l}^i \right |^{p}$; 
  \item \textbf{Attention Sum} (sum of the \kaiqi{activation} values) 
  
  $F_{sum}(A^i_j)= \sum_{k=1}^{h^i}\sum_{l=1}^{w^i} {\left | a_{k,l}^i \right |}^{p}$. 
\end{enumerate}
%
From these three, we choose Attention Mean, $F_{mean}(A^i_j)$ with $p$ equal to $1$ as the indicator to identify and prune unimportant filters. 
%
Also, in contrast to the related works, our proposed attention mapping function is robust to the inputs, \kaiqi{including real data or arbitrary random vectors}; the attention values are calculated by only one batch of randomly chosen training data. 
See Section \ref{sec:ablation} for ablation studies on the choices of attention functions and the effect of data for evaluating attention values.
%

To the best of our knowledge, we are the first to study the effectiveness of using attention theories for model compression tasks such as structured pruning and solve the challenges by designing novel attention mapping functions that are effective for filter pruning.

\subsection{Adaptive Iterative Pruning}\label{sec:adaptive_pruning}

\begin{figure}[t]
    \begin{algorithm}[H]
        \centering
        \small
        \caption{Accuracy-guaranteed Adaptive Pruning}
        \begin{algorithmic}[1]
            \STATE \textbf{Input:} A converged uncompressed network and the target accuracy loss $AccLossTarget$
            \STATE \textbf{Output:} The smallest model meeting the accuracy target
            \STATE Initialize: $T=0.0, \lambda=0.01$.
            \FOR {pruning round $r$ ($r \geq 1$)}
                \STATE Prune the model using $T[r]$ (Refer to Lines 7 and 8 in Algorithm~\ref{alg:overall_pruning})
                \STATE Rewind weights and learning rate (Refer to Lines 9 and 10 in Algorithm~\ref{alg:overall_pruning})
                \STATE Train the pruned model, and evaluate its accuracy $Acc[r]$ (Refer to Lines 11 and 12 in Algorithm~\ref{alg:overall_pruning})
                \STATE Calculate the accuracy loss $AccLoss[r]$: \\$AccLoss[r] = Acc[0] - Acc[r]$
                \IF {$AccLoss[r] < AccLossTarget$}
                    \IF {the changes of model size are within 0.1\% for several rounds}
                        \STATE Terminate
                    \ELSE
                        \STATE $\lambda[r+1] = \lambda[r]$
                        \STATE $T[r+1] = T[r] + \lambda[r+1]$
                    \ENDIF
                \ELSE
                    \STATE Find the last acceptable round $k$
                    \IF {$k$ has been used to roll back for several times}
                        \STATE Mark $k$ as unacceptable
                        \STATE \kaiqi{Go to Step 17}
                    \ELSE
                        \STATE Roll back model weights to round $k$
                        \STATE $\lambda[r+1] = \lambda[r]/2.0^{(C+1)}$ (\kaiqi{$C$} is the number of times for rolling back to round $k$)
                        \STATE $T[r+1] = T[k] + \lambda[r+1]$
                    \ENDIF
                \ENDIF
            \ENDFOR
        \end{algorithmic}
    \label{alg:pruning_policy}
    \end{algorithm}
\end{figure}

\noindent Our approach to pruning is to automatically and efficiently generate a pruned model that meets the users' different objectives. Automatic pruning means that users do not have to figure out how to configure the pruning process. Efficient pruning means that the pruning process should produce the user-desired model as quickly as possible. 
Users' pruning objectives can vary depending on the usage scenarios: 
1) Accuracy-critical tasks, like those used by self-driving cars, have stringent accuracy requirements, which are critical for safety, but do not have strict limits on their computing and storage usages; 2) Memory-constrained tasks, like those deployed on microcontrollers, have very limited available memory to store the models but do not have strict accuracy requirements; and 3) Latency-sensitive tasks, like those employed by virtual assistants where timely responses are desirable but accuracy is not a hard constraint.

In order to achieve automatic and efficient pruning, we propose three adaptive pruning policies to provide 1) Accuracy-guaranteed pruning which produces the most resource-efficient model with the acceptable accuracy loss; 2) Memory-constrained pruning which generates the most accurate model within a given memory footprint; and 3) FLOPs-constrained pruning which creates the most accurate model within a given computational intensity.
Specifically, our adaptive pruning method automatically adjusts the global threshold ($T$) used in our iterative structured pruning algorithm (Algorithm~\ref{alg:overall_pruning}) to quickly find the model that meets the pruning objective.
Other objectives (e.g., limiting a model's energy consumption) as well as multi-objective optimization can also be readily supported (See Section~\ref{sec:discussion}).
We take Accuracy-guaranteed Adaptive Pruning, described in Algorithm~\ref{alg:pruning_policy}, as an example to show the procedure of adaptive pruning. 
Algorithm~\ref{alg:pruning_policy} is a specific version of Algorithm~\ref{alg:overall_pruning} with the acceptable accuracy loss set as the pruning target.
Other versions with memory and FLOPs targets are included in Appendix~\ref{sec:appendix_adaptive_pruning}.

%
In the algorithm, $T$ controls the aggressiveness of pruning, and $\lambda$ determines the increment of $T$ at each pruning round. Pruning starts conservatively, with $T$ initialized to $0$, so that only completely useless filters that cannot capture any features are pruned. After each round, if the model accuracy loss is below the target accuracy loss, it is considered ``acceptable'', and the algorithm increases the aggressiveness of pruning by incrementing $T$ by $\lambda$, with $\lambda$ initialized to 0.01. 
%
%
%
As pruning becomes increasingly aggressive, the accuracy eventually drops below the target in a certain round which is considered ``unacceptable''. When this happens, our algorithm rolls back the model weights and pruning threshold to the last acceptable round where the accuracy loss is within the target, and restarts the pruning from there but more conservatively---it increases the threshold more slowly by cutting the $\lambda$ value by half.
If this still does not lead to an acceptable round, the algorithm cuts $\lambda$ by half again and restarts again.
If after several trials, the accuracy loss is still not acceptable, the algorithm rolls back even further and restarts from an earlier round. 
%
The rationale behind this adaptive algorithm is that the aggressiveness of pruning should accelerate when the model is far from the pruning target and decelerate when it is close to the target.

Note that the value of $\lambda$ continuously decreases as the algorithm gets close to the target, which guarantees the convergence of the pruned model: 
when $\lambda$ becomes sufficiently small, $T$ does not grow much anymore, which means no more filters are pruned and the model converges. See Section~\ref{sec:ablation} for an example. The algorithm terminates when the changes of model size is within 0.1\% for several rounds. 
%

The proposed algorithms can automatically---the only two parameters $T$ and $\lambda$ are automatically tuned---generate pruned models that meet diverse user requirements in model accuracy, size, and speed. Moreover, by making LTH-based iterative pruning adaptive, the algorithms can automatically find the best models that meet user requirements. Compared to related works which require time-consuming manual tuning of pruning parameters, AAP is the first to achieve these goals which are crucial to the practical use of model pruning for diverse real-world scenarios.

\subsection{Layer-aware Threshold Adjustment}
\label{sec:layer-aware}

While adapting the global pruning threshold using the above discussed policies, our pruning method further considers the difference in each layer's contribution to model size and complexity and uses differentiated layer-specific thresholds to prune the layers. 
As shown in Figure~\ref{fig:conv2d}, in terms of the contribution to model size, the number of parameters of layer $i$ can be estimated as $N^i = n^{(i-1)} \times k^i \times k^i \times n^i$; in terms of the contribution to computational complexity, the number of FLOPs of layer $i$ can be estimated as $F^i = 2 \times h^i \times w^i \times N^i$.
A layer that contributes more to the model's size or FLOPs is more likely to have redundant filters to prune without affecting the model's accuracy.

Therefore, to effectively prune a model while maintaining its accuracy, we need to treat each layer differently at each round of the iterative pruning process based on its current contributions to the model size and complexity.
Specifically, our adaptive pruning method calculates a weight for each layer based on its contribution and then uses this weight to adjust the current global threshold and derive a local threshold for the layer. 
If the goal is to reduce model size, the weight is calculated as each layer's number of remaining parameters $N^i[r]$ divided by the model's total number of remaining parameters $N^{Total}[r]$: $w^i[r] = \frac{N^i[r]}{N^{Total}[r]}$, where $r$ is the pruning round.
If the goal is to reduce model computational complexity, the weight is calculated as each layer's remaining FLOPs $F^i[r]$ divided by the model's total remaining FLOPs $F^{Total}[r]$: $w^i[r] = \frac{F^i[r]}{F^{Total}[r]}$. Then the threshold of layer $i$ is calculated as: $T^i[r] = T[r] \times w^i[r]$.
These layer-specific thresholds are then used to prune the layers in the current pruning round; they replace the global threshold $T[r]$ used to prune filters in Line 8 of Algorithm~\ref{alg:overall_pruning}.

Compared to the related works which often use a single threshold to prune parameters for the entire network \cite{han2015learning, zhao2019variational}, AAP's layer-specific thresholds allow it to generate better pruned models, and these thresholds are also fully automatically tuned.

\section{Evaluation}\label{sec:evaluation}

\begin{table}[t]
\centering
\scriptsize
\caption{\footnotesize{Implementation details}}
\vspace{-10pt}
\label{tab:implementation_details}
\setlength{\tabcolsep}{3.0pt}
\begin{tabular}{c|c|cc|cc}
\toprule
\multirow{2}{*}{Dataset}                                                  & \multirow{2}{*}{Model} & \multicolumn{2}{c|}{Learning Rate Schdeluder}                   & \multirow{2}{*}{\begin{tabular}[c]{@{}c@{}}Training \\ Epochs\end{tabular}} & \multirow{2}{*}{\begin{tabular}[c]{@{}c@{}}Weight \\ decay\end{tabular}} \\
                                                                          &                        & Initial LR & \begin{tabular}[c]{@{}c@{}}Decay \\ Epochs\end{tabular} &                                                                             &                                                                          \\ \hline
\multirow{2}{*}{MNIST}                                                    & LeNet-5                & 0.1                                                              & N/A                                                     & 100                                                                         & 0.0                                                                      \\
                                                                          & LeNet-300-100          & 0.0012                                                           & N/A                                                     & 100                                                                         & 0.0                                                                      \\ \hline
\multirow{7}{*}{CIFAR-10}                                                 & ResNet-56              & 0.1                                                              & {[}91, 136{]}                                   & 182                                                                         & 2.00E-04                                                                 \\
                                                                          & ResNet-50              & 0.1                                                              & {[}91, 136{]}                                   & 182                                                                         & 2.00E-04                                                                 \\
                                                                          & ShuffleNet             & 0.1                                                              & {[}60, 120, 160{]}                              & 200                                                                         & 4.00E-05                                                                 \\
                                                                          & MobileNet-V2           & 0.1                                                              & {[}150, 225{]}                                  & 300                                                                         & 4.00E-05                                                                 \\
                                                                          & VGG-16                 & 0.05                                                             & {[}150, 180, 210{]}                             & 240                                                                         & 5.00E-04                                                                 \\
                                                                          & VGG-19                 & 0.05                                                             & {[}150, 180, 210{]}                             & 240                                                                         & 5.00E-04                                                                 \\
                                                                          & LeNet-5                & 0.0002                                                           & N/A                                                     & 24                                                                          & 1.00E-04                                                                 \\ \hline
\multirow{2}{*}{\begin{tabular}[c]{@{}c@{}}Tiny-\\ ImageNet\end{tabular}} & ResNet-101             & 0.1                                                              & {[}150, 225{]}                                  & 300                                                                         & 2.00E-04                                                                 \\
                                                                          & VGG-19                 & 0.1                                                              & {[}150, 225{]}                                  & 300                                                                         & 2.00E-04                                                                 \\ \hline
ImageNet                                                                  & ResNet-50              & 0.256                                                            & {[}30, 60, 80{]}                                & 90                                                                          & 1.00E-04                                                                 \\ 
\bottomrule
\end{tabular}
\end{table}

\begin{table}[t]
\centering
\scriptsize
\caption{\footnotesize{Results from ResNet-56 and ResNet-50 on CIFAR-10. \kaiqi{For the Acc. ↓ (\%) column, a negative value means an increase in accuracy. The baseline Top-1 accuracy of ResNet-56 and ResNet-50 is 92.84\% and 91.83\%, respectively.}}}
\vspace{-10pt}
\label{tab:resnet_cifar10}
\setlength{\tabcolsep}{5.0pt}
\begin{tabular}{p{1.0cm}llllll}
\toprule
\multicolumn{1}{l}{Model}    & Target                                                                     & \begin{tabular}[c]{@{}c@{}}Target\\  Level\end{tabular} & Method          & \begin{tabular}[c]{@{}c@{}}Acc. \\ ↓ (\%)\end{tabular} & \begin{tabular}[c]{@{}c@{}}Params. \\ ↓ (\%)\end{tabular} & \begin{tabular}[c]{@{}c@{}}FLOPs. \\ ↓ (\%)\end{tabular} \\ \hline
\multirow{34}{*}{ResNet-56}  & \multirow{12}{*}{\begin{tabular}[c]{@{}c@{}}Acc. \\ ↓ (\%)\end{tabular}}   & \multirow{7}{*}{0\%}                                    & SCOP            & 0.06                                                   & 56.30                                                     & 56.00                                                    \\
                             &                                                                            &                                                         & HRank           & 0.09                                                   & 42.40                                                     & 50.00                                                    \\
                             &                                                                            &                                                         & SWP             & 0.03                                                   & 42.60                                                     & 43.60                                                    \\
                             &                                                                            &                                                         & ILP             & 0.00                                                   & 13.04                                                     & -                                                        \\
                             &                                                                            &                                                         & NSP             & -0.03                                                  & -                                                         & 47.00                                                    \\
                             &                                                                            &                                                         & EagleEye        & -1.40                                                   & -                                                         & 50.41                                                    \\
                             &                                                                            &                                                         & \textbf{AAP-P} & -0.33                                         & \textbf{79.11}                                            & \textbf{56.97}                                               \\
                             &                                                                            &                                                         & \textbf{AAP-F} & -0.08                                         & \textbf{65.78}                                                & \textbf{70.13}                                           \\ \cline{3-7} 
                             &                                                                            & \multirow{5}{*}{1\%}                                    & CP             & 1.00                                                   & 50.00                                                     & -                                                        \\
                             &                                                                            &                                                         & ILP             & 1.00                                                   & 41.18                                                     & -                                                        \\
                             &                                                                            &                                                         & GAL             & 0.52                                                   & 44.80                                                     & 48.50                                                    \\
                             &                                                                            &                                                         & \textbf{AAP-P} & 0.86                                          & \textbf{88.23}                                            & \textbf{70.09}                                               \\
                             &                                                                            &                                                         & \textbf{AAP-F} & 0.77                                          & \textbf{78.69}                                                & \textbf{81.19}                                           \\ \cline{2-7} 
                             & \multirow{7}{*}{\begin{tabular}[c]{@{}c@{}}Params. \\ ↓ (\%)\end{tabular}} & \multirow{4}{*}{70\%}                                   & DCP             & -0.01                                                  & 70.30                                                     & -                                                    \\
                             &                                                                            &                                                         & GBN             & 0.03                                                   & 66.70                                                     & -                                                    \\
                             &                                                                            &                                                         & HRank           & 2.38                                                   & 68.10                                                     & -                                                    \\
                             &                                                                            &                                                         & \textbf{AAP}   & \textbf{-0.60}                                          & 71.57                                            & -                                               \\ \cline{3-7} 
                             &                                                                            & \multirow{3}{*}{50\%}                                                        & SFP             & 1.33                                                   & 50.60                                                     & -                                                    \\
                             &                                                                            &                                                         & FPGM            & 0.10                                                   & 50.60                                                     & -                                                    \\
                             &                                                                            &                                                         & \textbf{AAP}   & \textbf{-1.06}                                         & 53.89                                            & -                                               \\ \cline{2-7} 
                             & \multirow{11}{*}{\begin{tabular}[c]{@{}c@{}}FLOPs. \\ ↓ (\%)\end{tabular}}  & \multirow{2}{*}{75\%}                                   & HRank           & 2.38                                                   & -                                                     & 74.10                                                    \\
                             &                                                                            &                                                         & \textbf{AAP}   & \textbf{0.97}                                          & -                                                & 75.97                                           \\ \cline{3-7} 
                             &                                                                            & \multirow{3}{*}{70\%}                                   & DeepHoyer              & 2.54                                                   & -                                                         & 71.00                                                    \\
                             &                                                                            &                                                         & NSP             & 1.17                                                   & -                                                         & 71.00                                                    \\
                             &                                                                            &                                                         & \textbf{AAP}   & \textbf{0.30}                                           & -                                                & 71.44                                           \\ \cline{3-7} 
                             &                                                                            & \multirow{6}{*}{55\%}                                   & CP              & 1.00                                                      & -                                                         & 50.00                                                    \\
                             &                                                                            &                                                         & SFP             & 1.33                                                   & -                                                     & 52.60                                                    \\
                             &                                                                            &                                                         & FPGM            & 0.10                                                   & -                                                     & 52.60                                                    \\
                             &                                                                            &                                                         & AMC             & 0.90                                                   & -                                                         & 50.00                                                    \\
                             &                                                                            &                                                         & SCP             & 0.46                                                   & -                                                         & 51.50                                                    \\
                             &                                                                            &                                                         & \textbf{AAP}   & \textbf{-0.63}                                         & -                                                & 52.92                                           \\ \hline
\multirow{2}{*}{ResNet-50}   & \multirow{2}{*}{\begin{tabular}[c]{@{}c@{}}Params. \\ ↓ (\%)\end{tabular}} & \multirow{2}{*}{60\%}                                   & AMC             & -0.02                                                  & 60.00                                                     & -                                                        \\
                             &                                                                            &                                                         & \textbf{AAP}   & \textbf{-0.86}                                         & 64.81                                            & -                                               \\
\bottomrule
\end{tabular}
\end{table}

\begin{table}[t]
\centering
\scriptsize
\caption{\footnotesize{Results from VGG, LetNet, MobileNet, and ShuffleNet models on CIFAR-10. \kaiqi{For the Acc. ↓ (\%) column, a negative value means an increase in accuracy.
The baseline Top-1 accuracy of VGG-16, VGG-19, MobileNet-V2, ShuffleNet, and LeNet-5 are 93.64\%, 93.90\%, 94.46\%, 93.28\%, and 69.67\%, respectively.}}}
\vspace{-10pt}
\label{tab:other_models_cifar10}
\setlength{\tabcolsep}{4.7pt}
\begin{tabular}{p{1.0cm}llllll}
\toprule
\multicolumn{1}{l}{Model}    & Target                                                                     & \begin{tabular}[c]{@{}c@{}}Target\\  Level\end{tabular} & Method          & \begin{tabular}[c]{@{}c@{}}Acc. \\ ↓ (\%)\end{tabular} & \begin{tabular}[c]{@{}c@{}}Params. \\ ↓ (\%)\end{tabular} & \begin{tabular}[c]{@{}c@{}}FLOPs. \\ ↓ (\%)\end{tabular} \\ \hline
\multirow{15}{*}{VGG-16}      & \multirow{5}{*}{\begin{tabular}[c]{@{}c@{}}Acc. \\ ↓ (\%)\end{tabular}}    & \multirow{5}{*}{0\%}                                    & PFS             & -0.19                                                  & -                                                        & 50.00                                                        \\
                             &                                                                            &                                                         & VCNNP           & 0.07                                                   & -                                                      & 60.90                                                        \\
                             &                                                                            &                                                         & Hinge           & 0.43                                                   & -                                                     & 39.07                                                        \\
                             &                                                                            &                                                         & HRank           & 0.53                                                   & -                                                      & 53.60                                                        \\
                             &                                                                            &                                                         & \textbf{AAP-F}   & -0.16                                         & \textbf{72.85}                                            & \textbf{61.17}                                           \\ \cline{2-7} 

                          & \multirow{8}{*}{\begin{tabular}[c]{@{}l@{}}Params. \\ ↓ (\%)\end{tabular}} & \multirow{4}{*}{70\%}                                   & IMP-Refill              & 0.10                                                    & 67.00                                                        & -                                                        \\
                                                                                   &                                                                            &                                                         & IMP-Refill+             & 0.63                                                   & 70.00                                                        & -                                                        \\
                                                                                   &                                                                            &                                                         & IMP-Regroup             & 0.10                                                    & 69.00                                                        & -                                                        \\
                                                                                   &                                                                            &                                                         & \textbf{AAP}                     & \textbf{-0.27}                                                  & 70.04                                                     & -                                                        \\ \cline{3-7} 
                                                                                   &                                                                            & \multirow{4}{*}{80\%}                                   & IMP-Refill              & 0.55                                                   & 80.00                                                        & -                                                        \\
                                                                                   &                                                                            &                                                         & IMP-Refill+             & -                                                    & 80.00                                                        & -                                                        \\
                                                                                   &                                                                            &                                                         & IMP-Regroup             & -0.05                                                  & 80.00                                                        & -                                                        \\
                                                                                   &                                                                            &                                                         & \textbf{AAP}                     & \textbf{-0.09}                                                  & 81.21                                                     & -                                                        \\ \hline
\multirow{7}{*}{VGG-19}      & \multirow{5}{*}{\begin{tabular}[c]{@{}c@{}}Acc. \\ ↓ (\%)\end{tabular}}    & \multirow{5}{*}{0\%}                                    & EigenDamage     & 0.19                                                   & 78.18                                                     & 37.13                                                    \\
                             &                                                                            &                                                         & NN Slimming     & 1.33                                                   & 80.07                                                     & 42.65                                                    \\
                                                          &                                                                            &                                                         & PFS     & -0.31                                                   & -                                                     & 52.00                                                    \\
                             &                                                                            &                                                         & \textbf{AAP-P}   & -0.26                                         & \textbf{85.99}                                            & \textbf{56.22}                                           \\  
                             &                                                                            &                                                         & \textbf{AAP-F}   & -0.03                                         & \textbf{87.63}                                            & \textbf{61.31}                                           \\ \cline{2-7} 

                             & \multirow{2}{*}{\begin{tabular}[c]{@{}c@{}}FLOPs. \\ ↓ (\%)\end{tabular}}  & \multirow{2}{*}{85\%}                                   & EigenDamage     & 1.88                                                   & -                                                         & 86.51                                                    \\
                             &                                                                            &                                                         & \textbf{AAP}   & \textbf{1.81}                                          & -                                                & 89.02                                           \\ \hline

\multirow{6}{*}{\begin{tabular}[c]{@{}c@{}}MobileNet \\ -V2 \end{tabular}} & \multirow{6}{*}{\begin{tabular}[c]{@{}c@{}}Acc. \\ ↓ (\%)\end{tabular}}    & \multirow{6}{*}{0\%}                                    & GDP             & -0.26                                                  & -                                                     & 46.22                                                        \\
                                 &                 &           & MDP             & -0.12                                                  & -                                                     & 28.71                                                        \\
                             &                        &                                    & SCOP            & 0.24                                                   & -                                                     & 40.30                                                        \\
                             &                                                                            &                                                         & DMC             & -0.26                                                  & -                                                        & 40.00                                                        \\
                             &                                                                            &                                                         & \textbf{AAP-P}   & -0.28                                          & \textbf{79.68}                                            & \textbf{55.67}                                           \\ 
                             &                                                                            &                                                         & \textbf{AAP-F}   & -0.26                                          & \textbf{76.79}                                            & \textbf{58.99}                                           \\ 

                             \hline 

\multirow{2}{*}{ShuffleNet}   & \multirow{2}{*}{\begin{tabular}[c]{@{}c@{}}Acc. \\ ↓ (\%)\end{tabular}} & \multirow{2}{*}{0\%}                                   & QP             & 0.31                                                  & 28.57                                                     & -                                                        \\
                             &                                                                            &                                                         & \textbf{AAP-P}   & 0.19                                         & \textbf{50.87}                                            & \textbf{26.67}                                               \\ \hline
                             
\multirow{2}{*}{LeNet-5}   & \multirow{2}{*}{\begin{tabular}[c]{@{}c@{}}Params. \\ ↓ (\%)\end{tabular}} & \multirow{2}{*}{90\%}                                   & ILP             & 10.24                                                  & 89.60                                                     & -                                                        \\
                             &                                                                            &                                                         & \textbf{AAP}   & \textbf{1.85}                                         & 90.38                                            & -                                               \\ 

\bottomrule
\end{tabular}
\end{table}

\begin{table}[t]
\scriptsize
\centering
\caption{\footnotesize{Results from ResNet-50 on ImageNet, and VGG-19 and ResNet-101 on Tiny-ImageNet. \kaiqi{For the Acc. ↓ (\%) column, a negative value means an increase in accuracy.
The baseline Top-1 accuracy of ResNet-50 (ImageNet), VGG-19 (Tiny-ImageNet), and ResNet-101 (Tiny-ImageNet) are 75.06\%, 59.64\%, and 52.93\%, respectively.}}} 
\vspace{-10pt}
\label{tab:resnet50_imagenet}
\setlength{\tabcolsep}{4.7pt}
\begin{tabular}{p{1.0cm}llllll}
\toprule
\multicolumn{1}{l}{Model}    & Target                                                                     & \begin{tabular}[c]{@{}c@{}}Target\\  Level\end{tabular} & Method          & \begin{tabular}[c]{@{}c@{}}Acc. \\ ↓ (\%)\end{tabular} & \begin{tabular}[c]{@{}c@{}}Params. \\ ↓ (\%)\end{tabular} & \begin{tabular}[c]{@{}c@{}}FLOPs. \\ ↓ (\%)\end{tabular} \\ \hline
\multirow{13}{*}{ResNet-50} & \multirow{6}{*}{\begin{tabular}[c]{@{}c@{}}Acc. \\ ↓ (\%)\end{tabular}} & \multirow{3}{*}{0\%}  & PFP-A         & 0.22           & 18.10          & 10.80          \\
                            &                                  &                       & \textbf{AAP-P} & -0.12 & \textbf{24.55} & \textbf{11.26}  \\
                            &                                  &                       & \textbf{AAP-F} & 0.18  & \textbf{24.09} & \textbf{11.62} \\ \cline{3-7} 
                            &                                  & \multirow{4}{*}{1\%}  & SSS-41        & 0.68           & 0.78           & 15.06          \\
                            &                                  &                       & ILP           & 1.00           & 11.50          & -          \\
                            &                                  &                       & \textbf{AAP-P} & 0.64  & \textbf{30.39} & \textbf{22.70} \\
                            &                                  &                       & \textbf{AAP-F} & 0.83  & \textbf{29.26} & \textbf{32.26} \\ \cline{2-7} 
                            & \multirow{5}{*}{\begin{tabular}[c]{@{}c@{}}Params. \\ ↓ (\%)\end{tabular}}  & \multirow{3}{*}{40\%} & SSS-26          & 4.30           & 38.82          & -          \\
                            &                                  &                       & Hrank         & 1.17           & 36.70          & -          \\
                            &                                  &                       & \textbf{AAP} & \textbf{1.69}  & 40.85 & -     \\ \cline{3-7} 
                            &                                  & \multirow{2}{*}{30\%} & PFP-B         & 0.92           & 30.10          & -          \\
                            &                                  &                       & \textbf{AAP} & \textbf{0.80}  & 31.19 & -     \\ \cline{2-7} 
                            & \multirow{2}{*}{\begin{tabular}[c]{@{}c@{}}FLOPs. \\ ↓ (\%)\end{tabular}}   & \multirow{2}{*}{30\%} & SSS-32        & 1.94           & -          & 31.08          \\
                            &                                  &                       & \textbf{AAP} & \textbf{0.57}  & -     & 31.88 \\ \cline{1-7} 
\multirow{5}{*}{VGG-19} & \multirow{3}{*}{\begin{tabular}[c]{@{}c@{}}Acc. \\ ↓ (\%)\end{tabular}}   & \multirow{2}{*}{3\%}       & EigenDamage     & 3.36          & 61.87           & 66.21             \\
                            &                                 &                       & \textbf{AAP-P} & 2.75 & \textbf{73.69} & \textbf{79.87} \\
                            &                                 &                       & \textbf{AAP-F} & 2.61 & \textbf{72.58} & \textbf{78.97} \\
                        \cline{3-7} 
                        & \multirow{2}{*}{\begin{tabular}[c]{@{}c@{}}Params. \\ ↓ (\%)\end{tabular}}   & \multirow{3}{*}{60\%}       & NN Slimming     & 10.66          & 60.14           & -             \\
                            &                                 &                       & \textbf{AAP} & \textbf{-0.34} & 60.21 & - \\ \hline
                                                                                                                                                
\multirow{6}{*}{ResNet-101}   & \multirow{6}{*}{\begin{tabular}[c]{@{}c@{}}Acc. \\ ↓ (\%)\end{tabular}} & \multirow{5}{*}{0\%} & NN Slimming  & 1.36  & 75.00    & 75.00   \\
                            &                                                                            &             & GAL   & 0.50  & 45.00 & {76.00}                                               \\ 
                            &                                                                            &             & DHP   & 0.01  & 50.00 & {75.00}                                               \\ 
                            &                                                                            &             & ACTD   & -0.44  & 51.00 & {75.00}                                               \\ 
                            &                                                                            &             & \textbf{AAP-P}   & -0.57  & \textbf{92.72} & \textbf{76.58}                                               \\ 

\bottomrule
\end{tabular}
\end{table}

\begin{table}[t]
\centering
\scriptsize
\caption{\footnotesize{\kaiqi{Results from LetNet-5 and LetNet-300-100 on MNIST. For the Acc. ↓ (\%) column, a negative value means an increase in accuracy.
The baseline Top-1 accuracy of LetNet-5 and LeNet-300-100 are 99.11\% and 97.34\%, respectively.}}}
\vspace{-10pt}
\label{tab:resnet_mnist}
\begin{tabular}{llllll}
\toprule
Model                                                                              & Target                                                                     & \begin{tabular}[c]{@{}l@{}}Target \\ Level\end{tabular} & Method                  & \begin{tabular}[c]{@{}l@{}}Acc. \\ ↓ (\%)\end{tabular} & \begin{tabular}[c]{@{}l@{}}Params. \\ ↓ (\%)\end{tabular}  \\ \hline
\multirow{2}{*}{\begin{tabular}[c]{@{}l@{}}LeNet-300-100\end{tabular}}   & \multirow{2}{*}{\begin{tabular}[c]{@{}l@{}}Params. \\ ↓ (\%)\end{tabular}} & \multirow{2}{*}{90\%}                                   & PFP & 0.41                                                   & 84.32                                                                                                            \\
                                                                                   &                                                                            &                                                         & \textbf{AAP}                     & \textbf{0.38}                                                   & 89.94                                                                                                           \\ \hline
\multirow{6}{*}{\begin{tabular}[c]{@{}l@{}}LeNet-5\end{tabular}}         & \multirow{6}{*}{\begin{tabular}[c]{@{}l@{}}Params. \\ ↓ (\%)\end{tabular}} & \multirow{6}{*}{99\%}                                   & PFP & 0.35                                                   & 92.37                                                                                                           \\
                                                                                   &                                                                            &                                                         & AP+Coreset-K            & 0.00                                                   & 99.39                                                                                                        \\
                                                                                   &                                                                            &                                                         & AP+Coreset-S            & 0.01                                                   & 99.48                                                                                                            \\
                                                                                   &                                                                            &                                                         & AP+Coreset-A            & 0.01                                                   & 99.48                                                                                                          \\
                                                                                   &                                                                            &                                                         & ADMM-NN-S               & 0.20                                                   & 98.86                                                                                                          \\
                                                                                   &                                                                            &                                                         & \textbf{AAP}                     & \textbf{-0.01}                                                  & 99.34                                                                                                           \\ \bottomrule
\end{tabular}
\end{table}

We did an extensive evaluation on diverse models (ResNet, VGG, MobileNet, LeNet and ShuffleNet) and datasets (\kaiqi{MNIST \cite{xiao2017fashion}}, CIFAR-10 \cite{krizhevsky2009learning}, Tiny-ImageNet \cite{le2015tiny} and ImageNet \cite{deng2009imagenet}), and provided a thorough comparison to SOTA works (discussed in Section \ref{sec:background} \kaiqi{and \ref{sec:methodology}}). 
For the proposed AAP, we consider different pruning policies: 1) meet accuracy target while minimizing the number of model parameters (\textit{AAP-P}) or FLOPs (\textit{AAP-F}); 2) meet parameter reduction target while minimizing accuracy loss (\textit{AAP}); and 3) meet FLOP reduction target while minimizing accuracy loss (\textit{AAP}).
Multi-objective, multi-constraint pruning with AAP is also considered (Section \ref{sec:discussion}).

\smallskip
\noindent \textbf{Implementation Details} 
%
%
%
%
\kaiqi{
We implemented AAP on PyTorch version 1.6.0 and conducted experiments on four Nvidia RTX 2080 GPUs.
The implementation details are present in Table~\ref{tab:implementation_details}. 
The learning rate decays with a factor of 0.1 at decay epochs. 
Nesterov SGD optimizer is used with a momentum of 0.9. 
The batch size is set to 256, 128, 256, and 64 for the models on MNIST, CIFAR-10, Tiny-ImageNet, and ImageNet, respectively.
Simple data augmentation (random crop and random horizontal flip) is used for all training images.}

\subsection{Results on CIFAR-10}\label{sec:results_on_cifar10}
In all cases targeting accuracy, model size, and compute intensity, the proposed method AAP significantly outperforms the recent related works. 
Table~\ref{tab:resnet_cifar10} shows the results from the widely used ResNet models on CIFAR-10. 
For example, for ResNet-56, without accuracy drop, AAP achieves the largest parameters reduction (79.11\%), outperforming the related works by 22.81\% to 66.07\%, and the largest FLOPs reduction (70.13\%), outperforming the related works by 14.13\% to 26.53\%.
With 70\% of parameters reduction, AAP achieves the smallest accuracy loss (-0.6\%), outperforming the related works by 0.59\% to 2.98\%.
Also, with 70\% of FLOPs reduction, AAP achieves the smallest accuracy loss (0.3\%), outperforming the related works by 0.87\% to 2.24\%. 
Note that the proposed AAP also produces a pruned model that reaches 0.6\% or 1.06\% higher accuracy than the original model but with only 28.43\% or 46.11\%, respectively, of the original parameters. Such small and accurate models are useful for many real-world applications.


%
Table~\ref{tab:other_models_cifar10} shows the results from other model architectures. 
\kaiqi{For example, on VGG-16, without any accuracy drop, AAP achieves the largest FLOPs reduction (61.17\%), outperforming the related works by 0.27\% to 22.1\%.}
AAP can also effectively compress models (MobileNet, ShuffleNet) that are already designed to be compact. 
\kaiqi{For example, on MobileNet-V2, without an accuracy drop, AAP achieves the largest FLOPs reduction (58.99\%), speeding up this already lightweight model by more than half and outperforming the related works by 12.77\% to 30.28\%.}




\subsection{Results on ImageNet and Tiny-ImageNet}

Table~\ref{tab:resnet50_imagenet} shows that AAP can also significantly outperform the SOTA methods for various models trained on ImageNet and Tiny-ImageNet. 
%
For example, on ResNet-50 with ImageNet, for the same level of accuracy loss, AAP reduces significantly more parameters (6.45\% to 29.61\% higher than the related works) and more FLOPs (0.82\% to 17.2\% higher than the related works);
For the same level of parameters or FLOPs reduction, AAP achieves the smallest accuracy loss, lower than the related works by 0.08\% to 2.61\%. 
On VGG-19 with Tiny-ImageNet, for the same level of parameters reduction, AAP achieves significantly lower accuracy loss than NN Slimming~\cite{liu2017learning} by 11\%. 
On ResNet-101 with Tiny-ImageNet, without accuracy loss, AAP achieves the highest parameters reduction (92.72\%), outperforming the related works by 17.72\% to 47.72\%.

\subsection{Results on MNIST}\label{sec:results_on_cifar10}
\kaiqi{
Table~\ref{tab:resnet_mnist} shows that AAP can also significantly outperform the related works for LeNet-5 and LetNet-300-100 trained on MNIST. 
On LeNet-5, with 99\% of parameters reduction, AAP achieves the smallest accuracy loss (-0.01\%), outperforming the related works by 0.01\% to 0.36\%.
Note that ADMM-NN-S~\cite{ma2021non} applies quantization after pruning, and AAP can also be further improved using quantization.
Even without quantization, with the same level of parameters reduction, AAP achieves a lower accuracy loss than ADMM-NN-S by 0.21\%.
On LeNet-300-100, with 90\% of parameters reduction, AAP achieves a lower accuracy loss than PFP~\cite{liebenwein2019provable} by 0.03\%.}

\subsection{Ablation Study}
\label{sec:ablation}

\begin{figure*}[t]
	\centering
	\begin{subfigure}{0.65\columnwidth}
		\centering
  		\includegraphics[width=4.5cm]{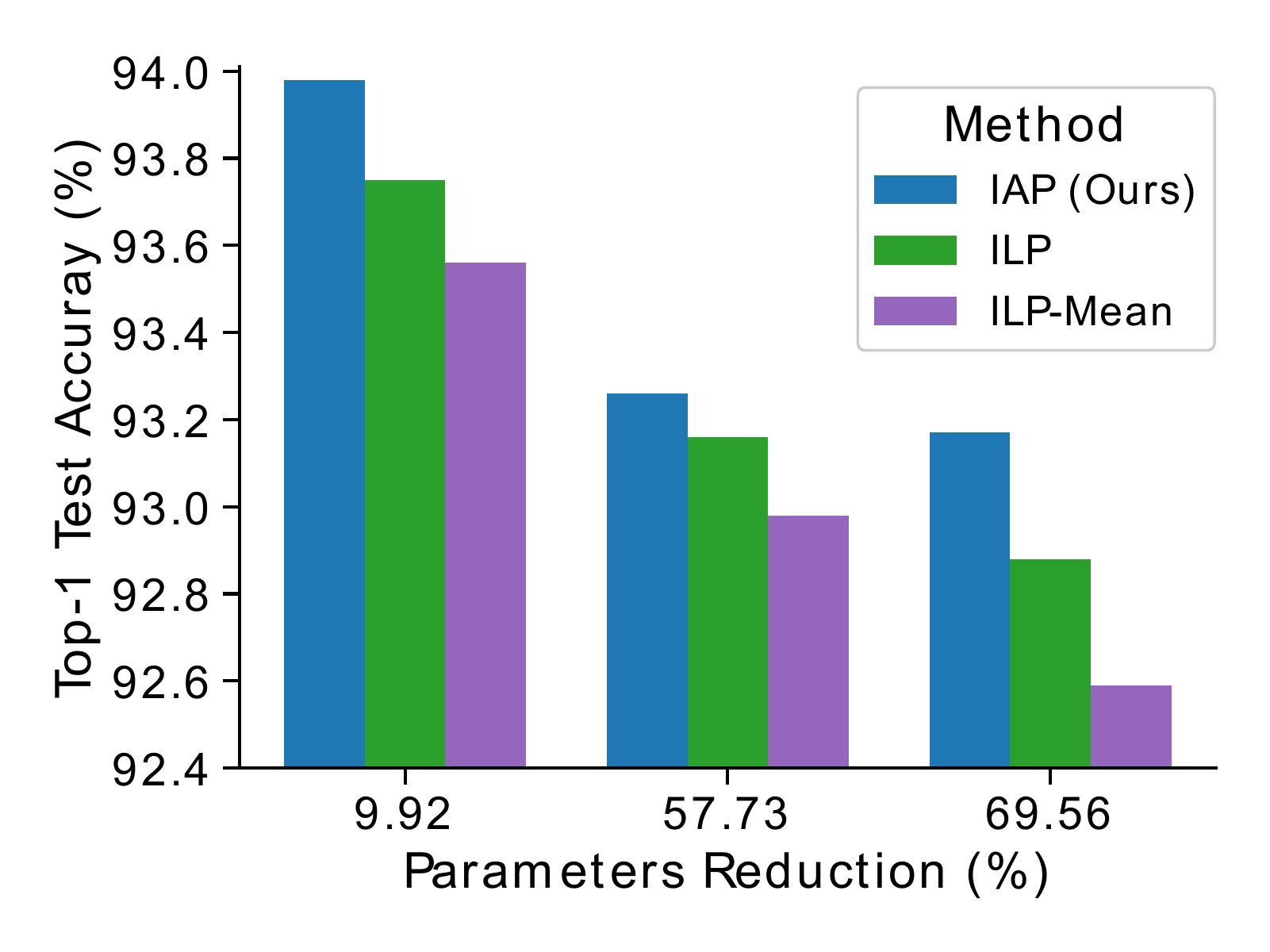}
		\caption{Effect of Attention Pruning}
		\label{fig:ablation_IAP}
	\end{subfigure}
	\begin{subfigure}{0.65\columnwidth}
		\centering
		\includegraphics[width=4.5cm]{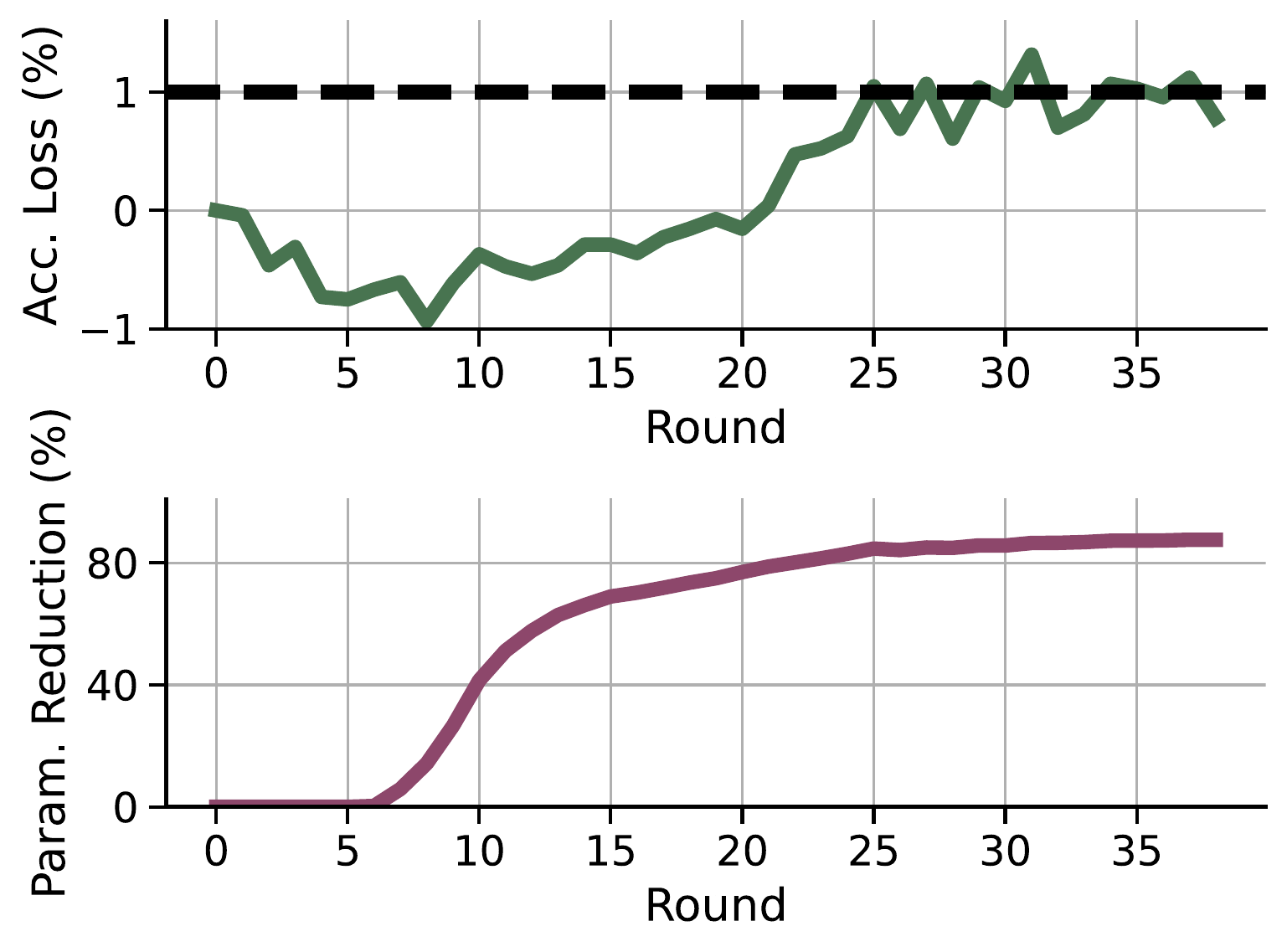}
		\caption{Effect of Adaptive Pruning}
		\label{fig:ablation_adaptive_pruning}
	\end{subfigure}
 	\begin{subfigure}{0.65\columnwidth}
		\centering
		\includegraphics[width=4.5cm]{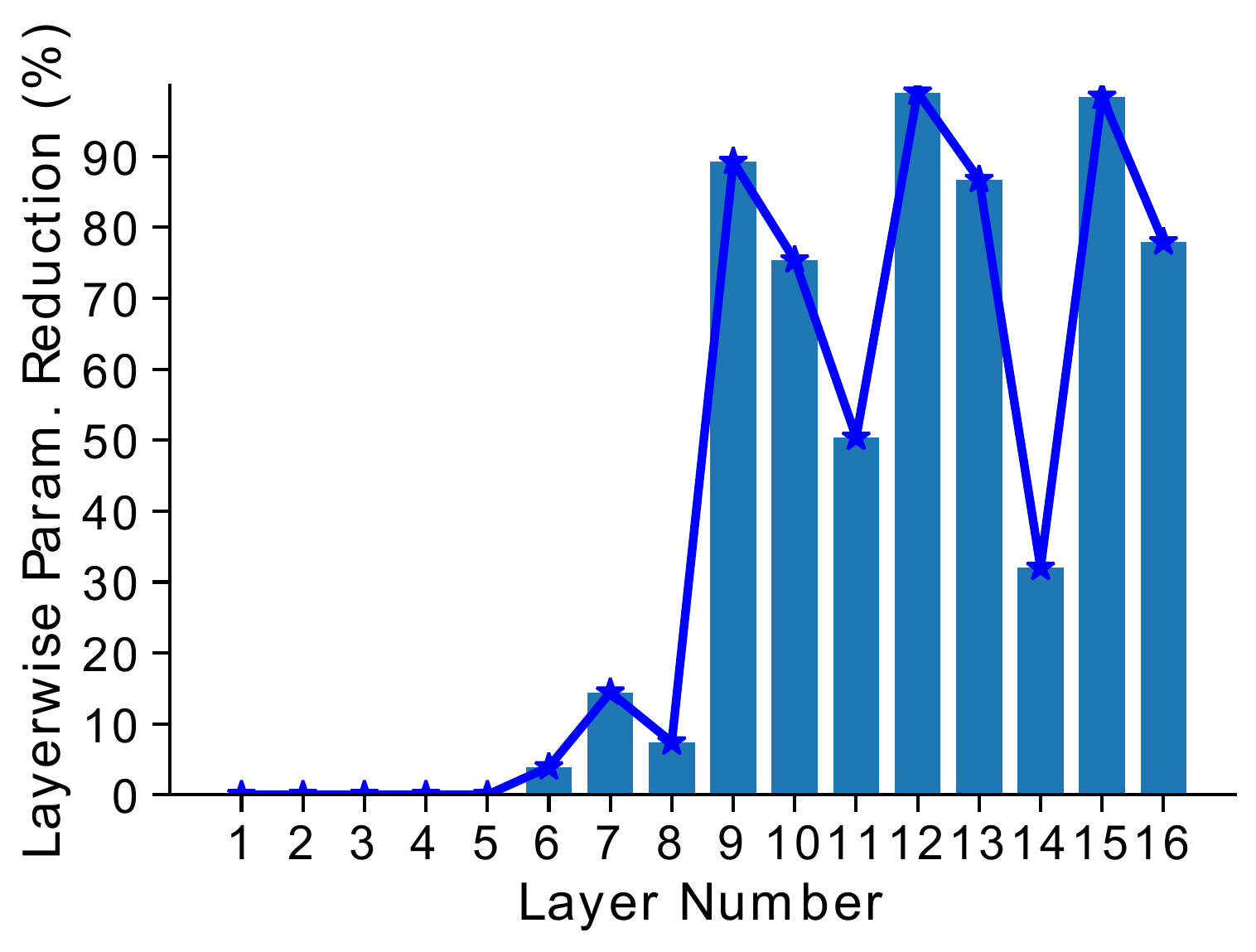}
		\caption{Layer-wise Sparsity}
		\label{fig:ablation_layerwise_sparsity}
	\end{subfigure}
    \caption{\footnotesize{(a) \kaiqi{Top-1 accuracy of VGG-16 iteratively pruned by the proposed attention pruning vs. L1-Norm based pruning on CIFAR-10. (b) Accuracy loss and parameter reduction over the pruning rounds as the pruning threshold is adapted following Algorithm~\ref{alg:pruning_policy}. (c) Layer-wise sparsity of a pruned VGG-19 on CIFAR-10.}}}
\end{figure*}

\begin{figure}[t]
	\centering
	\begin{subfigure}{0.47\columnwidth}
		\centering
		\includegraphics[width=4cm]{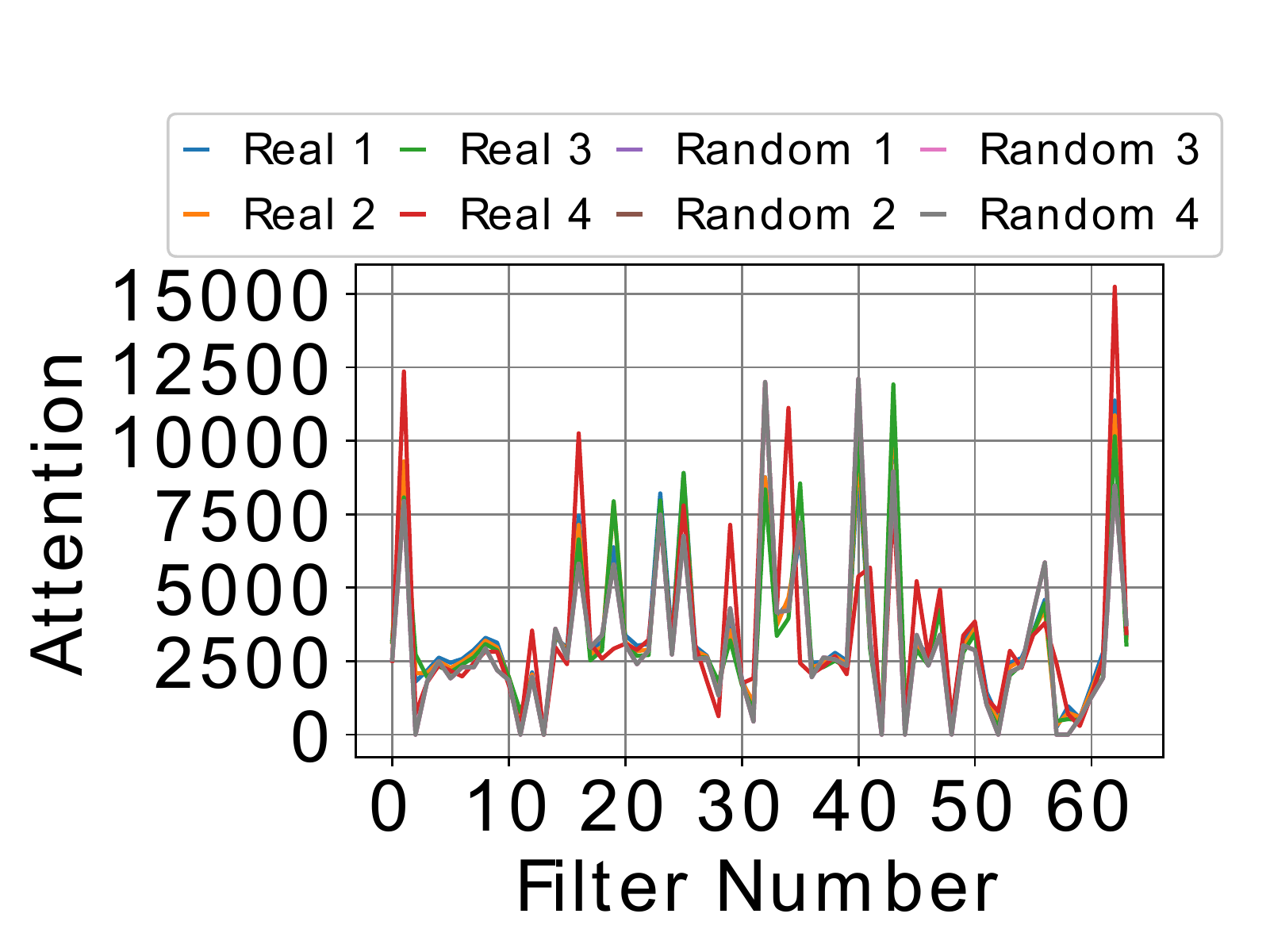}
		\caption{Conv1}
	\end{subfigure}
	\begin{subfigure}{0.47\columnwidth}
		\centering
		\includegraphics[width=4cm]{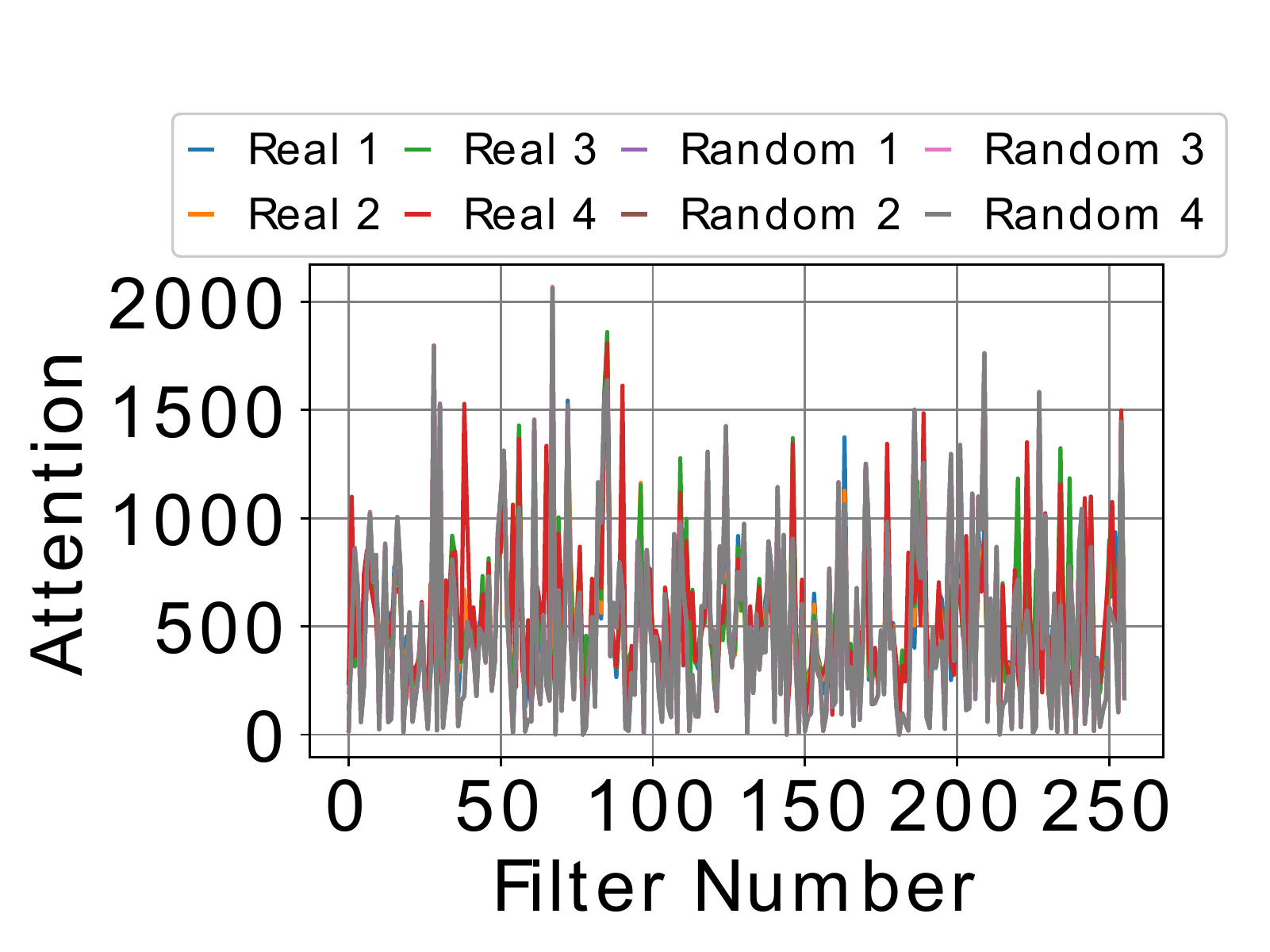}
		\caption{Group1.Block2.Conv3}
	\end{subfigure}
	\begin{subfigure}{0.47\columnwidth}
		\centering
		\includegraphics[width=4cm]{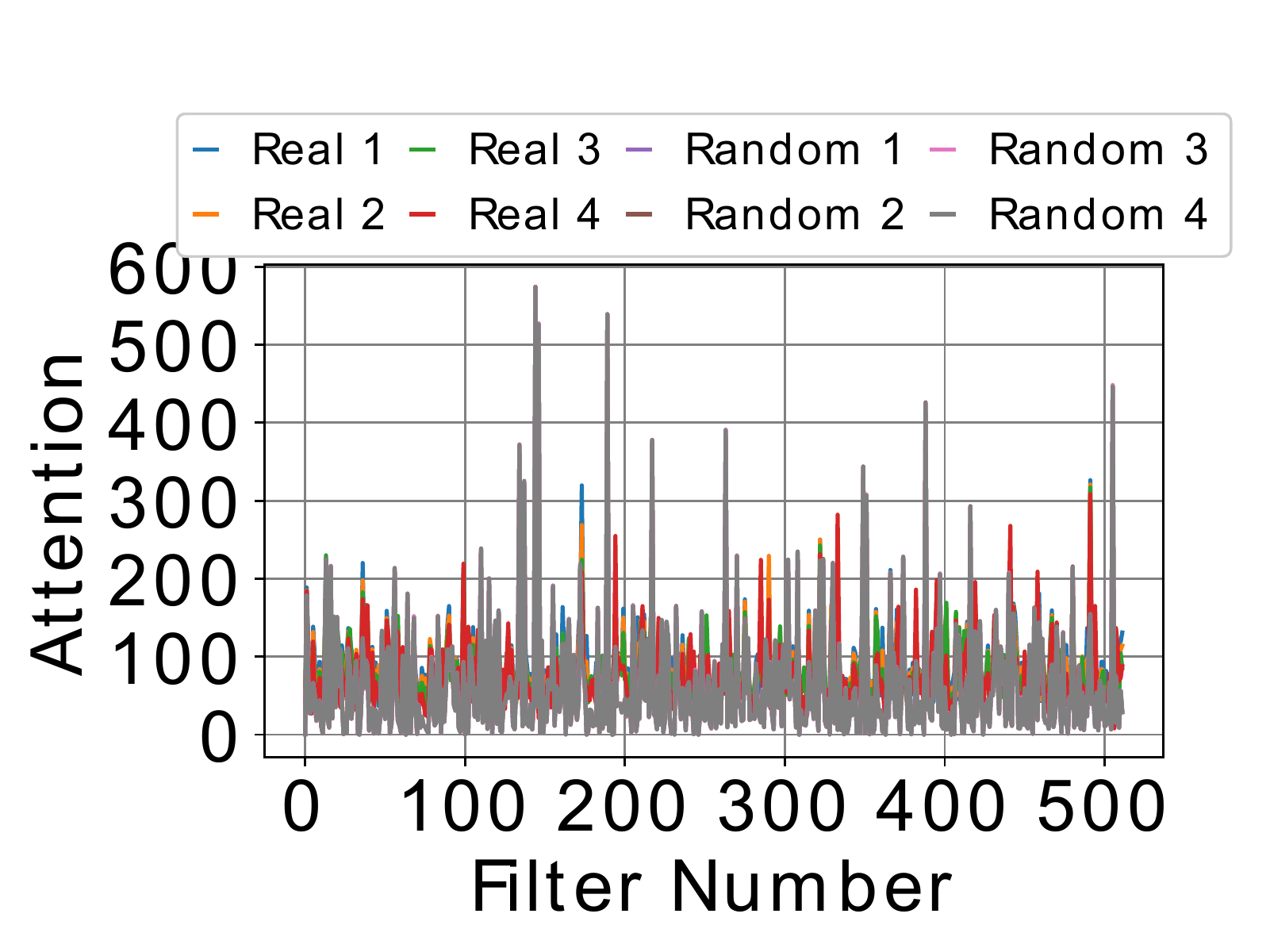}
		\caption{Group2.Block3.Conv3}
	\end{subfigure}
	\begin{subfigure}{0.47\columnwidth}
		\centering
		\includegraphics[width=4cm]{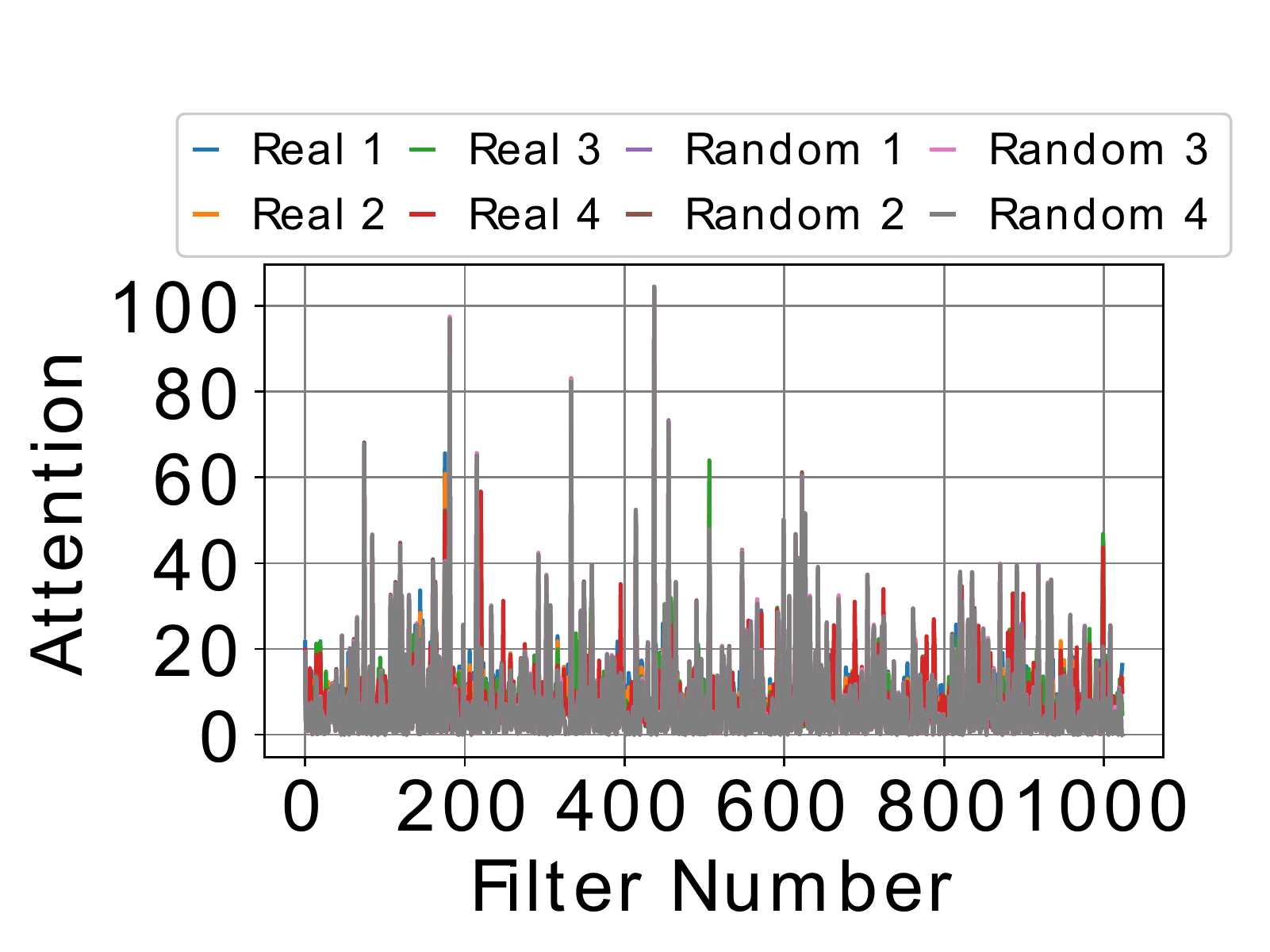}
		\caption{Group3.Block5.Conv3}
	\end{subfigure}
	\caption{{Attention values of filters from 4 convolution layers of ResNet-50 on ImageNet, given 8 different batches of inputs, including randomly chosen real images and arbitrary random vectors.}}
	\label{fig:attention_distribution}
\end{figure}

\noindent \textbf{The effect of attention-based iterative pruning.}
\kaiqi{We fix the percentage of filters that the pruning process removes in each round, which is non-adaptive iterative pruning, and compare using our proposed attention values (termed IAP) with 1) using the conventional weight values, e.g. the L1-Norm of filters (as in Iterative L1-Norm Pruning (ILP)~\cite{renda2020comparing}), and 2) using the L1-Norm of filters divided by the filter size (termed ILP-Mean) to choose which filters to remove.
Figure~\ref{fig:ablation_IAP} shows the Top-1 accuracy of VGG-16 with the parameters reduction of 9.92\%, 57.73\% and 69.56\% pruned by IAP, ILP, and ILP-Mean on CIFAR-10.
%
IAP leads to a higher accuracy than the others by 0.1\% to 0.58\%.} 



\smallskip
\noindent \textbf{The effect of attention mapping functions.}
Using VGG-16 on CIFAR-10 as an example, first, we analyze different types of attention mapping functions (discussed in Section~\ref{sec:filterPruning}), i.e., Attention Mean ($p=1$), Attention Sum ($p=1$), and Attention Max ($p=1$), when used in one-shot attention-based pruning. With a parameter reduction of 57.73\%, Attention Mean leads to the lowest top-1 accuracy loss, lower than Attention Sum and Attention Max by 0.25\% and 0.13\%, respectively. Then, we analyze Attention Mean with different values of $p$. When $p$ is set to $1$, it leads to the lowest accuracy loss: 0.38\% when $p=1$ vs 0.42\% when $p=2$ and 0.47\% when $p=4$. This confirms our choice of Attention Mean with $p=1$ in our evaluation.

\smallskip
\noindent \textbf{The effect of adaptive pruning.}
Using ResNet-56 on CIFAR-10 as an example, we show in Figure~\ref{fig:ablation_adaptive_pruning} how the adaptation of the pruning threshold (following Algorithm~\ref{alg:pruning_policy}) affects accuracy loss and parameters reduction over the pruning rounds. 
From Round 1 to Round 24, the accuracy loss of the pruned model is lower than the target accuracy loss (1\%), so the algorithm increases the pruning aggressiveness gradually by increasing the threshold. At Round 25, the accuracy loss exceeds the target, so the algorithm rolls back the model weights and the pruning threshold back to Round 24, and restarts the pruning from there more conservatively. The above process repeats until after Round 39, the model size converges, and the algorithm terminates at reducing 88.23\% of parameters.

\smallskip
\noindent \textbf{The effect of Layer-aware Threshold Adjustment.}
\kaiqi{AAP considers the importance of each layer by adjusting differentiated layer-specific thresholds. 
Figure~\ref{fig:ablation_layerwise_sparsity} shows the layer-wise sparsity of a pruned VGG-19 with a total parameters reduction of 85.99\% on CIFAR-10. 
More filters are pruned from higher layers than lower layers.} 

\smallskip
\noindent \textbf{The effect of inputs for evaluating attention values}.
The importance of each filter is evaluated by its attention value, which is calculated by one batch of randomly chosen training data after each pruning round (Line 7 in Algorithm~\ref{alg:overall_pruning}). The attention values are \textit{not sensitive} to the inputs because the model is converged from the training phase.
\kaiqi{As an example, Figure~\ref{fig:attention_distribution} shows the attention value of each filter of 4 convolution layers of ResNet-50 on ImageNet, given eight different batches of inputs, including randomly chosen real images and arbitrary random vectors.} 
The attention value of each filter is consistent across different batches. 

\subsection{Discussions}\label{sec:discussion}

\noindent \textbf{Comparison to related works.}
The above results validate that 1) the proposed \textit{activation-based attention pruning} is more effective than weight-magnitude-based pruning (e.g.,  EigenDamage \cite{wang2019eigendamage}, ILP \cite{renda2020comparing}, NSP \cite{zhuang2020neuron}) in finding unimportant filters;
%
%
%
2) the proposed \textit{activation-based attention pruning} is better than other activation-based pruning techniques that are based on different forms of activation feature maps, such as NN Slimming~\cite{liu2017learning}, HRank~\cite{lin2020hrank}, \kaiqi{AP+Coreset~\cite{dubey2018coreset} and PFP~\cite{liebenwein2019provable}};
%
%
%
and 3) the proposed \textit{adaptive pruning} can achieve better results than other automatic pruning methods (e.g., AMC~\cite{he2018amc} and GAL~\cite{lin2019towards}). 

%
%


\smallskip
\noindent \textbf{Inference speedup.}
The reduction in model complexity in FLOPs that AAP achieves does translate to real speedup in model inference. We conducted inference experiments using PyTorch framework on Raspberry Pi 3B+, a widely used \kaiqi{Internet-of-Things (IoT)} platform. 
%
Table~\ref{tab:inference} shows our method significantly improves the inference speed. 

\smallskip
\noindent \textbf{Extension to multi-objective optimization.}
Our method can be readily extended to support \textit{multiple objectives} and \textit{multiple constraints} for optimizing the pruned model in terms of accuracy, size, and/or speed simultaneously.
%
%
Table~\ref{tab:multi_optimization} shows two examples from pruning VGG-19 with CIFAR-10. In the first example, the objective is to minimize the accuracy loss under the constraints of 80\% reduction in \textit{both} FLOPs and parameters. In the second example, the objective is to maximize \textit{both} the FLOPs reduction and parameters reduction given no more than 1\% accuracy loss. 


\begin{table}[t]
\centering
\scriptsize
\caption{\footnotesize{Inference speedup on Raspberry Pi. \kaiqi{For each type of model, the two rows are two pruned models from the same uncompressed model with different levels of parameters reduction.}}}
\vspace{-10pt}
\label{tab:inference}
\begin{tabular}{lllll}
\toprule
Model and Dataset                                                                              & Target & Acc. ↓ (\%) & FLOPs. ↓ (\%) & Speedup \\ \hline
\multirow{2}{*}{\begin{tabular}[c]{@{}l@{}}ResNet-56\\ (CIFAR-10)\end{tabular}}      & \multirow{2}{*}{0\%}                                      & -0.10        & 33.77         & 1.20$\times$                                                          \\
                                                                                     &                                                         & -0.09       & 56.33         & 1.49$\times$                                                         \\ \hline
\multirow{2}{*}{\begin{tabular}[c]{@{}l@{}}VGG-16 \\ (CIFAR-10)\end{tabular}}        & \multirow{2}{*}{1\%}                                      & -0.17       & 59.51         & 2.13$\times$                                                         \\
                                                                                     &                                                         & 0.59        & 73.89         & 3.60$\times$                                                          \\ \hline
\multirow{2}{*}{\begin{tabular}[c]{@{}l@{}}ResNet-50\\ (Tiny-ImageNet)\end{tabular}} & \multirow{2}{*}{5\%}                                      & -3.92       & 35.06         & 1.01$\times$                                                         \\
                                                                                     &                                                         & -1.62       & 58.16         & 1.49$\times$                                                         \\ 
\bottomrule
\end{tabular}
\end{table}

\begin{table}[t]
\centering
\scriptsize
\caption{\footnotesize{Multi-objective optimization on VGG-19 (CIFAR-10).}}
\vspace{-10pt}
\label{tab:multi_optimization}
\begin{tabular}{lp{2.05cm}p{0.3cm}p{0.5cm}p{0.5cm}}
\toprule
Optimization Objectives                 & Constraints                            & \begin{tabular}[c]{@{}l@{}}Acc. \\ ↓ (\%)\end{tabular}            & \begin{tabular}[c]{@{}l@{}}Params. \\ ↓ (\%)\end{tabular}         & \begin{tabular}[c]{@{}l@{}}FLOPs \\ ↓ (\%)\end{tabular}          \\ \hline
\multirow{2}{*}{Minimize Accuracy Loss} & Params.  ↓  \textgreater 80\%     & \multirow{2}{*}{1.20}  & \multirow{2}{*}{83.11} & \multirow{2}{*}{80.60}  \\
                                        & and FLOPs  ↓  \textgreater 80\%       &                       &                        &                        \\ \hline
Maximize FLOP Reduction            & \multirow{2}{*}{Acc. ↓ \textless 1\%} & \multirow{2}{*}{0.15} & \multirow{2}{*}{89.11} & \multirow{2}{*}{62.51} \\
and Params. Reduction           &                                       &                       &                        &                        \\ 
\bottomrule
\end{tabular}
\end{table}

\section{Conclusions}\label{conclusion}



This paper proposes Automatic Attention Pruning (AAP), an adaptive, attention-based, structured pruning solution to automatically and efficiently generate small, accurate, and hardware-efficient models that meet diverse user requirements. We show that activation-based attention is a more precise indicator for identifying unimportant filters to prune than the commonly used weight magnitude value. We also offer an effective way to perform structured pruning in an adaptive process and find small and accurate sub-networks that are at the same time hardware efficient. Finally, we argue that automatic pruning is essential for pruning to be useful in practice, and propose an adaptive method that can automatically meet diverse user objectives in terms of model accuracy, size, and inference speed but without user intervention.
Our results confirm that our solution outperforms existing structured pruning approaches by a large margin.

\section{Acknowledgments}
We thank the anonymous reviewers for their feedback. 
This work is partly supported by National Science Foundation awards CNS-1955593 and OAC-2126291 and an Amazon Machine Learning Research Award.

\bibliographystyle{apalike}
\bibliography{reference}

\vfill\pagebreak
\appendix
\onecolumn
\section{Appendix}\label{sec:appendix}

In the supplementary materials, we first discuss the adaptive pruning policy with memory and FLOPs targets in Section~\ref{sec:appendix_adaptive_pruning}. 
Then we introduce the results of the additional experiments in Section~\ref{sec:appendix_experiments}.

\subsection{Adaptive Pruning Policy}\label{sec:appendix_adaptive_pruning}

\subsubsection{Memory-constrained Adaptive Pruning}\label{sec:appendix_memory_pruning}

The Memory-constrained Adaptive Pruning Algorithm is shown in Algorithm~\ref{alg:memory_pruning_policy}. 

\begin{algorithm}[H]
    \centering
    \small
    \caption{Memory-constrained Adaptive Pruning}
    \begin{algorithmic}[1]
        \STATE \textbf{Input:} Target Parameters Reduction $ParamTarget$
        \STATE \textbf{Output:} A small pruned model with an acceptable model size
        \STATE Initialize: $T=0.0, \lambda=0.01$.
        \FOR {pruning round $r$ ($r \geq 1$)}
            \STATE Prune the model using $T[r]$ 
            \STATE Rewind weights and the learning rate 
            \STATE Train the pruned model, and calculate its remaining number of parameters $Param[r]$ 
            
            \STATE Calculate the parameter reduction: $ParamRed[r]$: $ParamRed[r] = Param[0] - Param[r]$
            \IF {$ParamRed[r] < ParamTarget$}
                \IF {the changes of model size are within 0.1\% for several rounds} 
                    \STATE Terminate
                \ELSE
                    \STATE $\lambda[r+1] = \lambda[r]$
                    \STATE $T[r+1] = T[r] + \lambda[r+1]$
                \ENDIF
            \ELSE
                \STATE Find the last acceptable round $k$
                \IF {$k$ has been used to roll back for several times}
                    \STATE Mark $k$ as unacceptable
                    \STATE Go to Step 17
                \ELSE
                    \STATE Roll back model weights to round $k$
                    \STATE $\lambda[r+1] = \lambda[r]/2.0^{(C+1)}$ ($C$ is the number of times for rolling back to round $k$)
                    \STATE $T[r+1] = T[k] + \lambda[r+1]$
                \ENDIF
            \ENDIF
        \ENDFOR
    \end{algorithmic}
\label{alg:memory_pruning_policy}
\end{algorithm}

\vfill

\subsubsection{FLOPs-constrained Adaptive Pruning}\label{sec:appendix_flops_pruning}

The FLOPs-constrained Adaptive Pruning Algorithm is shown in Algorithm~\ref{alg:flops_pruning_policy}.
\begin{algorithm}[H]
    \centering
    \small
    \caption{FLOPs-constrained Adaptive Pruning}
    \begin{algorithmic}[1]
        \STATE \textbf{Input:} Target FLOPs Reduction $FLOPsTarget$
        \STATE \textbf{Output:} A small pruned model with an acceptable FLOPs
        \STATE Initialize: $T=0.0, \lambda=0.01$.
        \FOR {pruning round $r$ ($r \geq 1$)}
            \STATE Prune the model using $T[r]$ 
            \STATE Rewind weights and the learning rate 
            \STATE Train the pruned model, and calculate its remaining FLOPs $FLOPs[r]$ 
            
            \STATE Calculate the parameters reduction: $FLOPsRed[r]$: $FLOPsRed[r] = FLOPs[0] - FLOPs[r]$
            \IF {$FLOPsRed[r] < FLOPsTarget$}
                \IF {the changes of FLOPs are within 0.1\% for several rounds} 
                    \STATE Terminate
                \ELSE
                    \STATE $\lambda[r+1] = \lambda[r]$
                    \STATE $T[r+1] = T[r] + \lambda[r+1]$
                \ENDIF
            \ELSE
                \STATE Find the last acceptable round $k$
                \IF {$k$ has been used to roll back for several times}
                    \STATE Mark $k$ as unacceptable
                    \STATE Go to Step 17
                \ELSE
                    \STATE Roll back model weights to round $k$
                    \STATE $\lambda[r+1] = \lambda[r]/2.0^{(C+1)}$ ($C$ is the number of times for rolling back to round $k$)
                    \STATE $T[r+1] = T[k] + \lambda[r+1]$
                \ENDIF
            \ENDIF
        \ENDFOR
    \end{algorithmic}
\label{alg:flops_pruning_policy}
\end{algorithm}

\vfill

\newpage

\subsection{Additional Experiments}\label{sec:appendix_experiments}

\subsubsection{The Effect of Rewinding Epoch}\label{sec:appendix_rewinding}

\begin{figure*}[htp]
	\centering
	\begin{subfigure}{0.35\textwidth}
		\centering
  		\includegraphics[width=4.5cm]{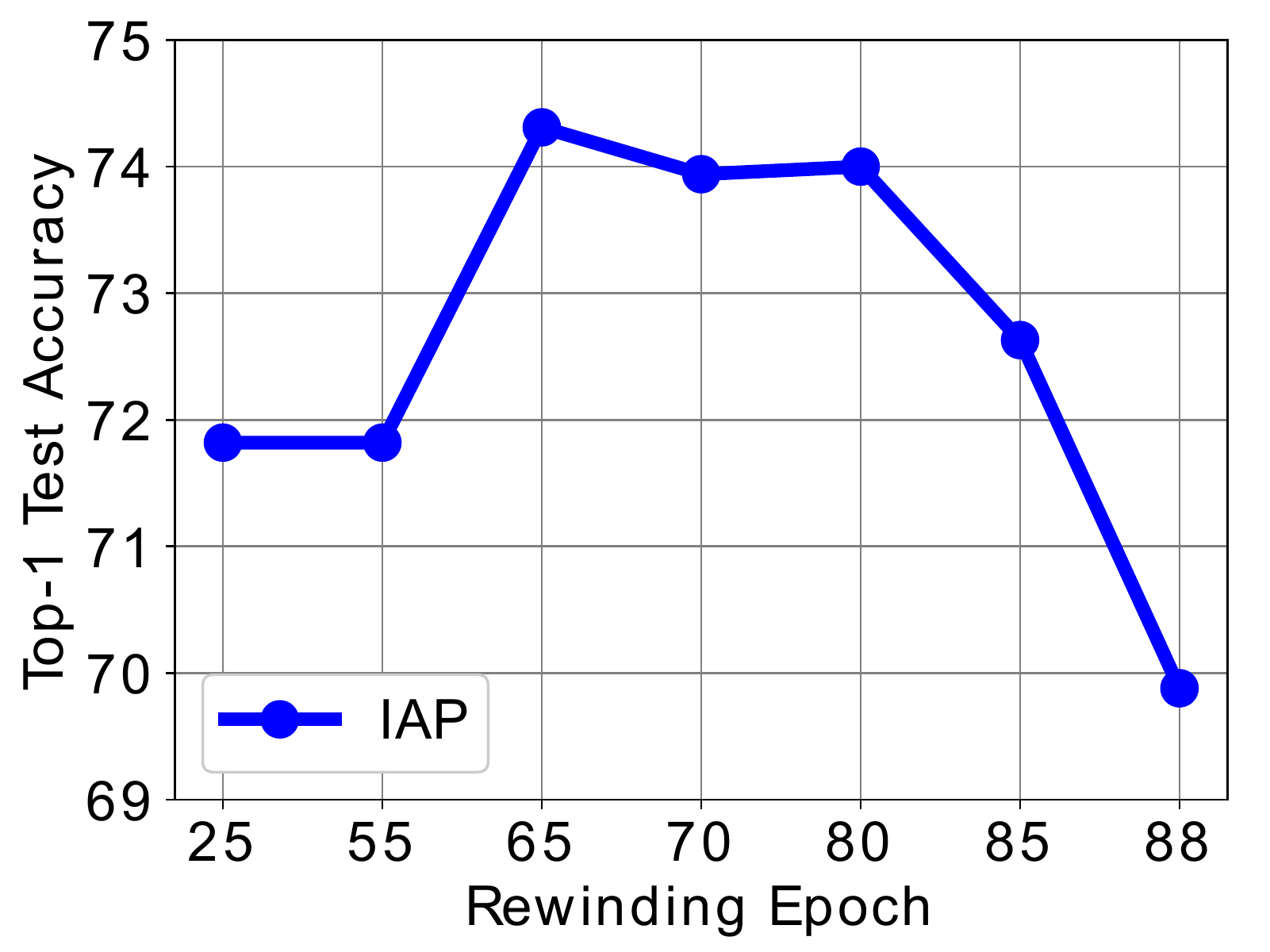}
		\caption{Top-1 Test Accuracy}
		\label{fig:ablation_accuracy}
	\end{subfigure}
	\begin{subfigure}{0.35\textwidth}
		\centering
  		\includegraphics[width=4.5cm]{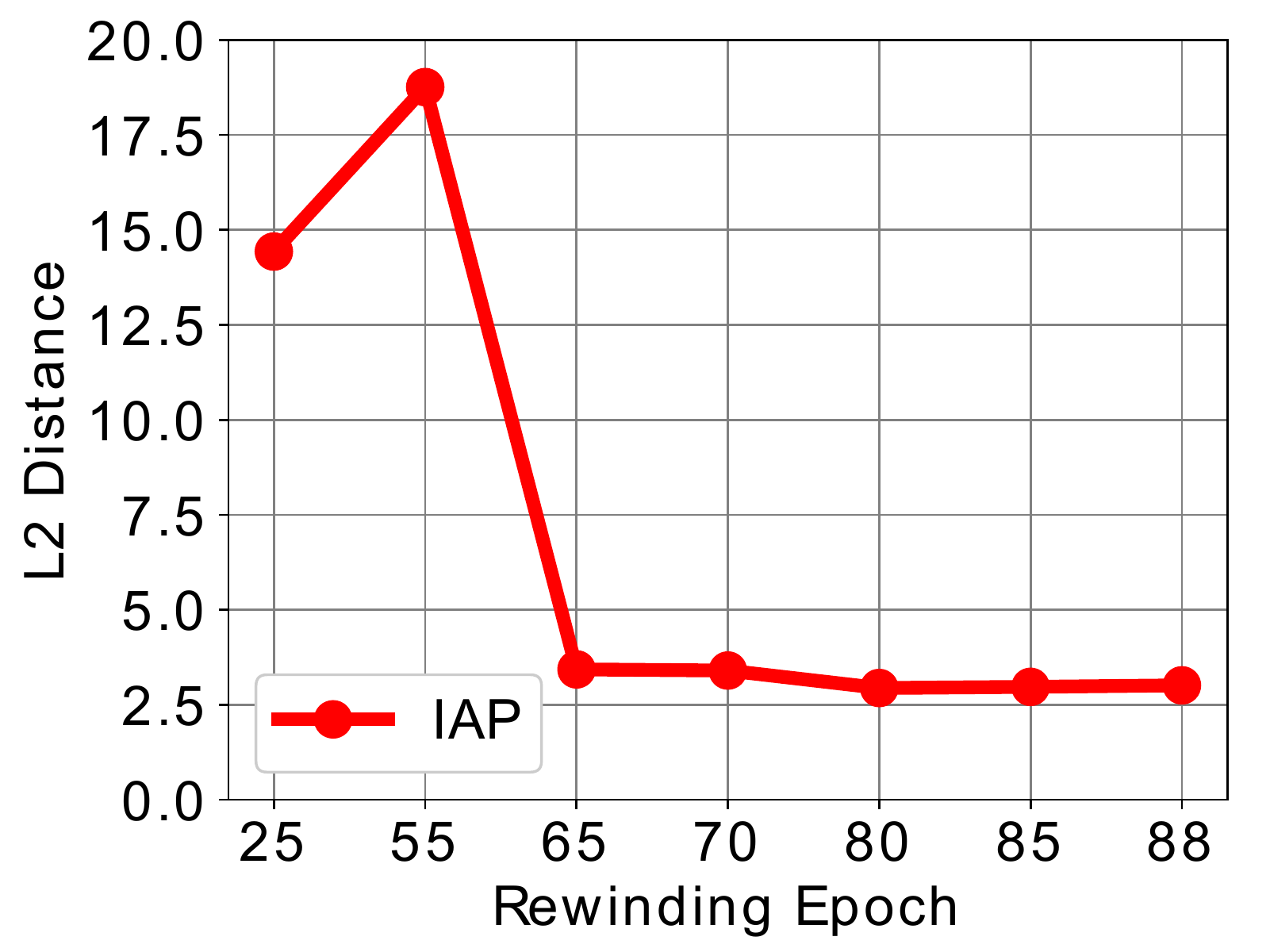}
		\caption{Stability to Pruning}
		\label{fig:ablation_stability}
	\end{subfigure}
	\caption{The effect of the rewinding epoch (x-axis) on  (a) Top-1 test accuracy, and (b) pruning stability, for pruned ResNet-50 with a parameter reduction of 16.26\% on ImageNet.}
\end{figure*}

To understand how the rewinding impacts the accuracy of the pruned models, we analyze \emph{stability to pruning}, which is defined as the L2 distance between the masked weights of the pruned network and the original network at the end of training.
We validate the observations that for deep networks, rewinding to very early stages is sub-optimal as the network has not learned considerably by then; and rewinding to very late training stages is also sub-optimal because there is not enough time to retrain. Specifically,
Figure~\ref{fig:ablation_accuracy} shows the Top-1 test accuracy of the pruned ResNet-50 with a parameter reduction of 16.26\% on ImageNet when the learning rate is rewound to different epochs, and Figure~\ref{fig:ablation_stability} shows the stability values at the corresponding rewinding epochs. We observe that there is a region, 65 to 80 epochs, where the resulting accuracy is high. We find that the L2 distance closely follows this pattern, showing a large distance for early training epochs and a small distance for later training epochs. 
Our findings show that rewinding to 75\%-90\% of training time leads to good accuracy.


\subsubsection{The Effect of Attentions}\label{sec:appendix_attention}

\begin{figure*}[htp]
	\centering
	\begin{subfigure}{0.33\linewidth}
		\centering
  		\includegraphics[width=2.0in]{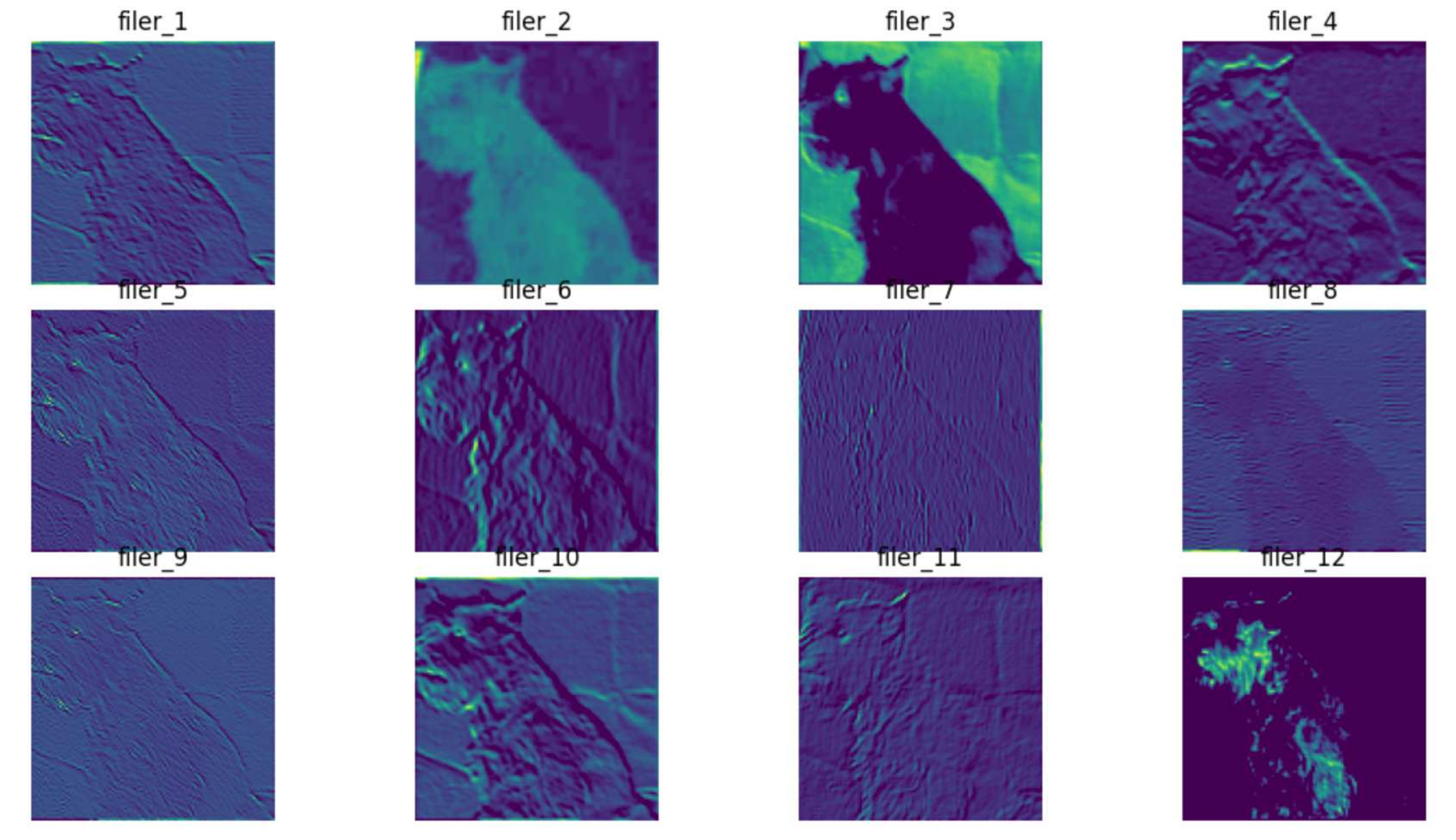}
		\caption{p=1}
	\end{subfigure}
	\begin{subfigure}{0.33\linewidth}
		\centering
  		\includegraphics[width=2.0in]{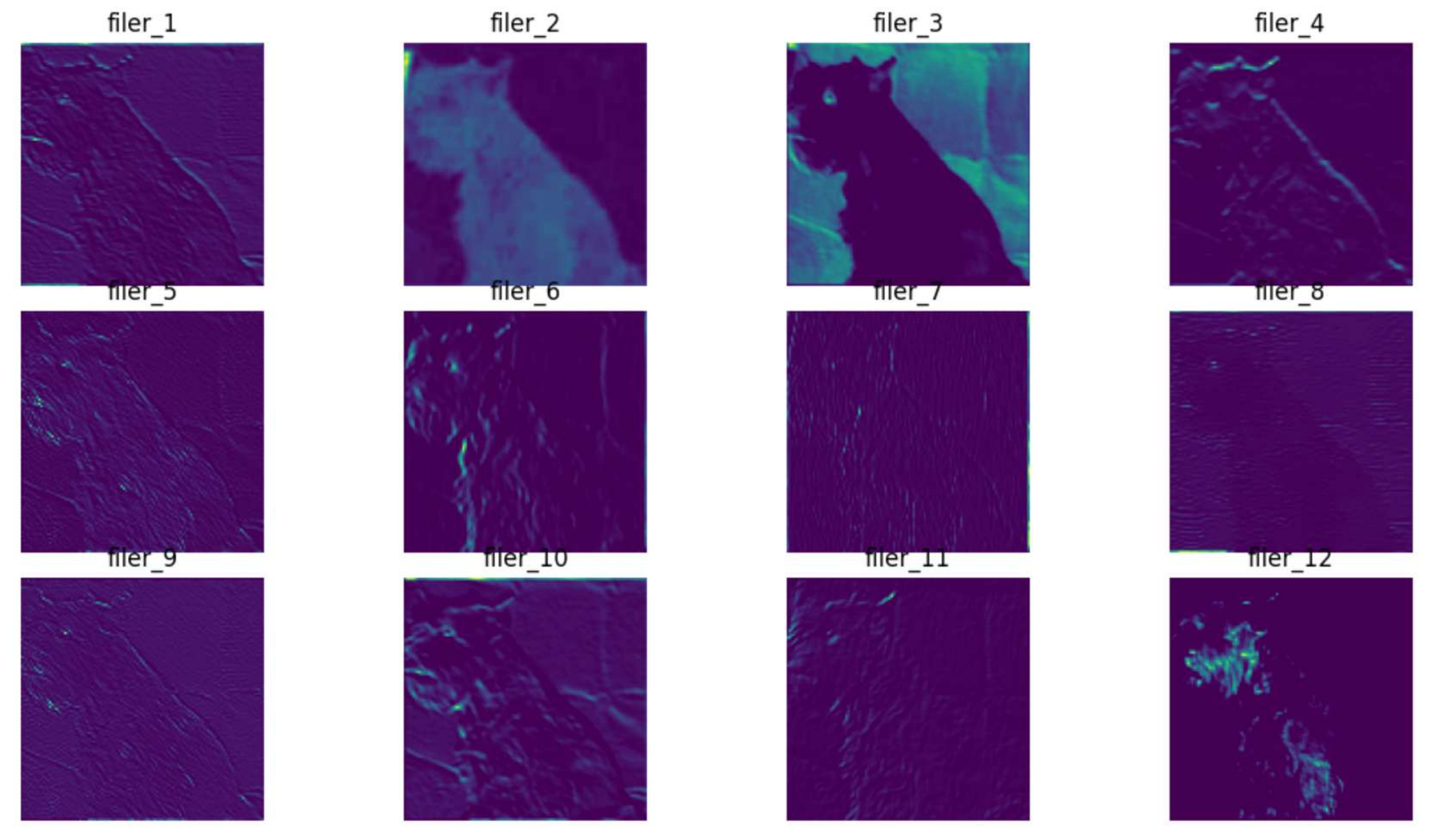}
		\caption{p=2}
	\end{subfigure}
 	\begin{subfigure}{0.33\linewidth}
		\centering
  		\includegraphics[width=2.0in]{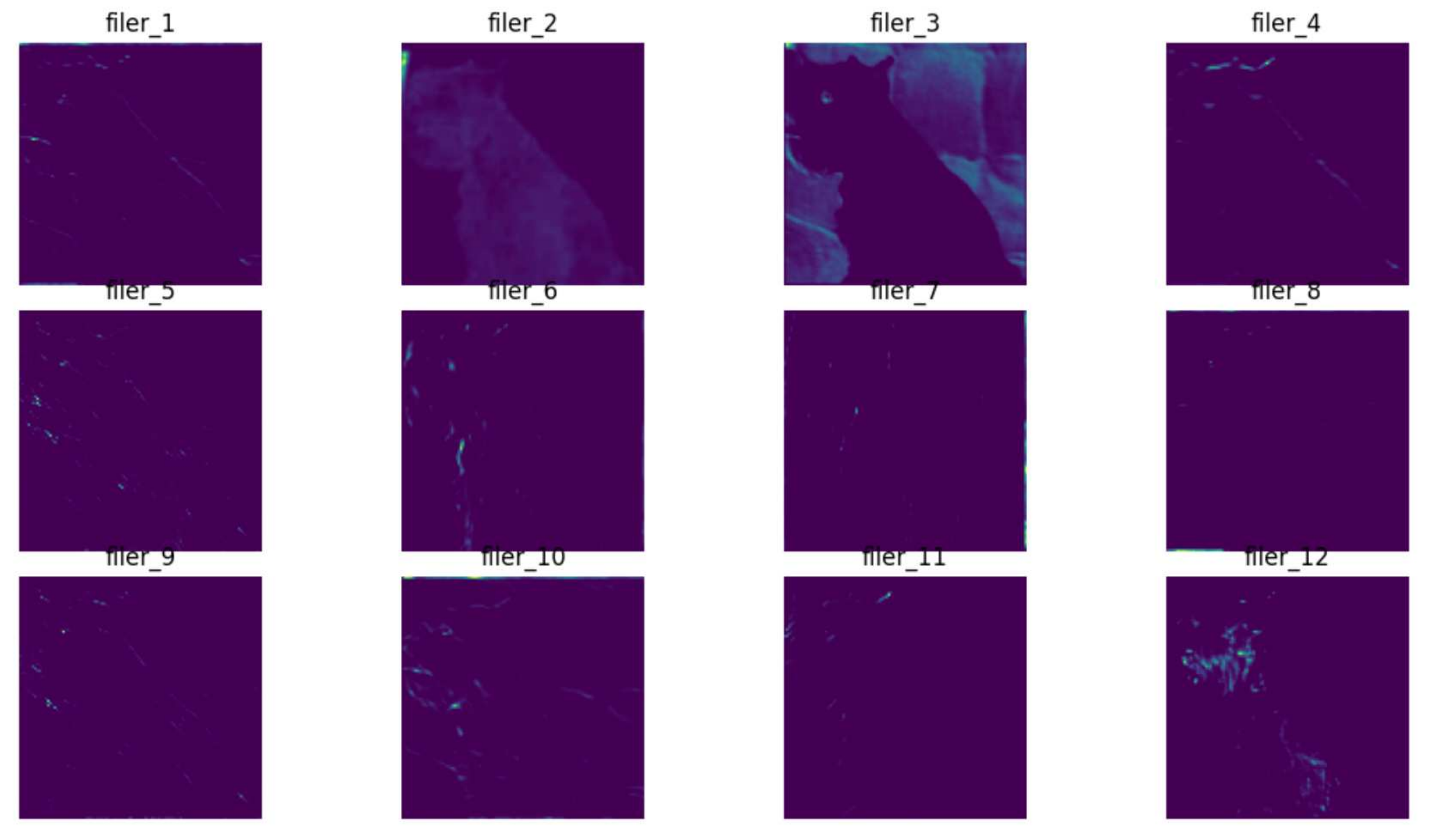}
		\caption{p=4}
	\end{subfigure}
	\caption{Attentions of each filter of the first convolution layer of ResNet-50 on ImageNet with different values of $p$ ($p = 1, 2, 4$).} 
 	\label{fig:attention}
\end{figure*}

Figure~\ref{fig:attention} shows the attention of each filter of the first convolution layer of ResNet-50 on ImageNet with different values of $p$ (p=1, 2, 4). The setting where $p$ is equal to 1 tends to be best since it promotes the effectiveness of the pruning by enabling the gap between the mean values of the useful and useless filters to be large.



\newpage
\subsubsection{Inference Speedup on CPU}
\begin{figure*}[htp]
	\centering
	\begin{subfigure}{0.4\textwidth}
		\centering
  		\includegraphics[width=6.0cm]{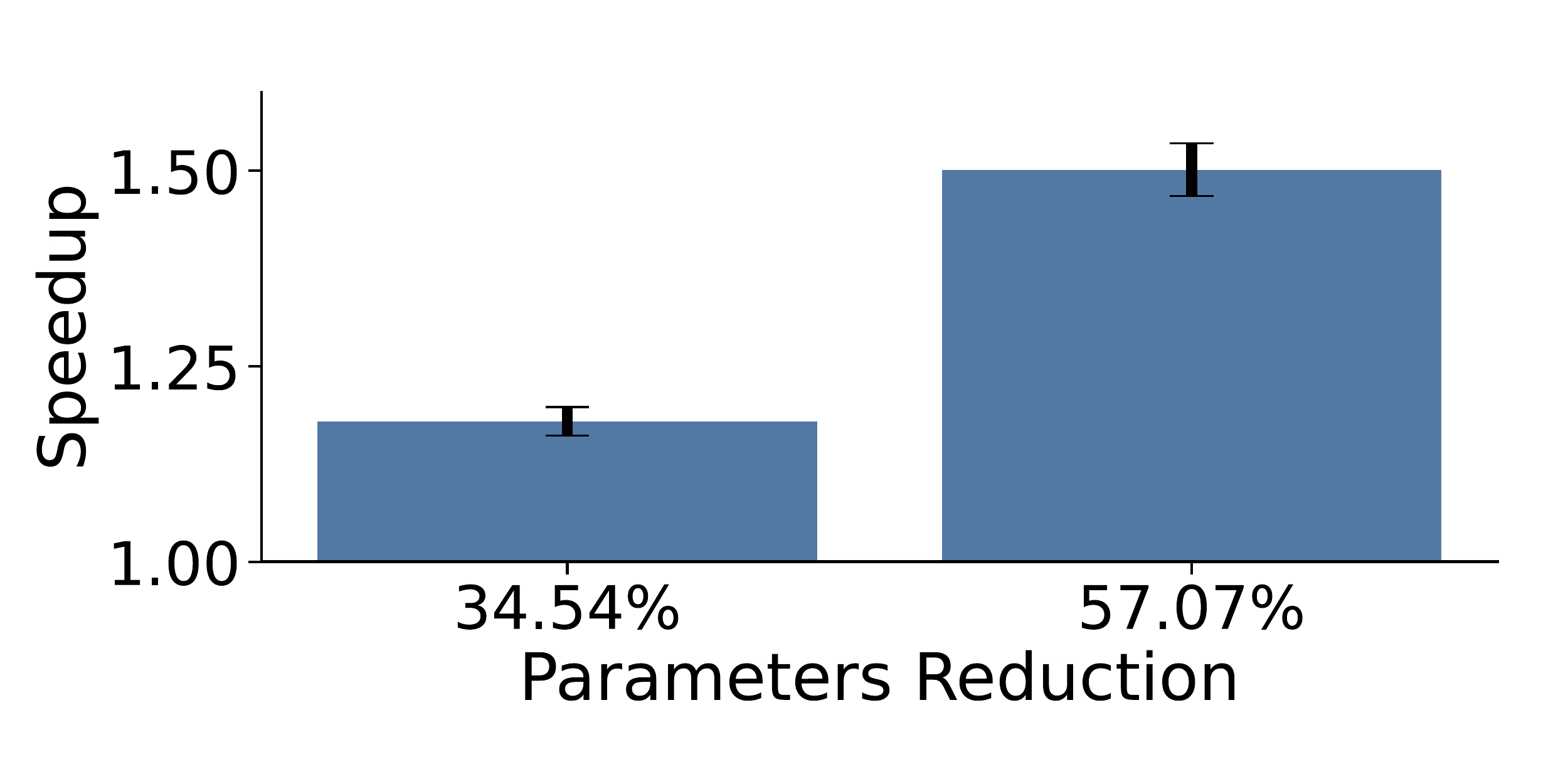}
        \caption{Speedup}
	\end{subfigure}
	\begin{subfigure}{0.4\textwidth}
		\centering
  		\includegraphics[width=6.0cm]{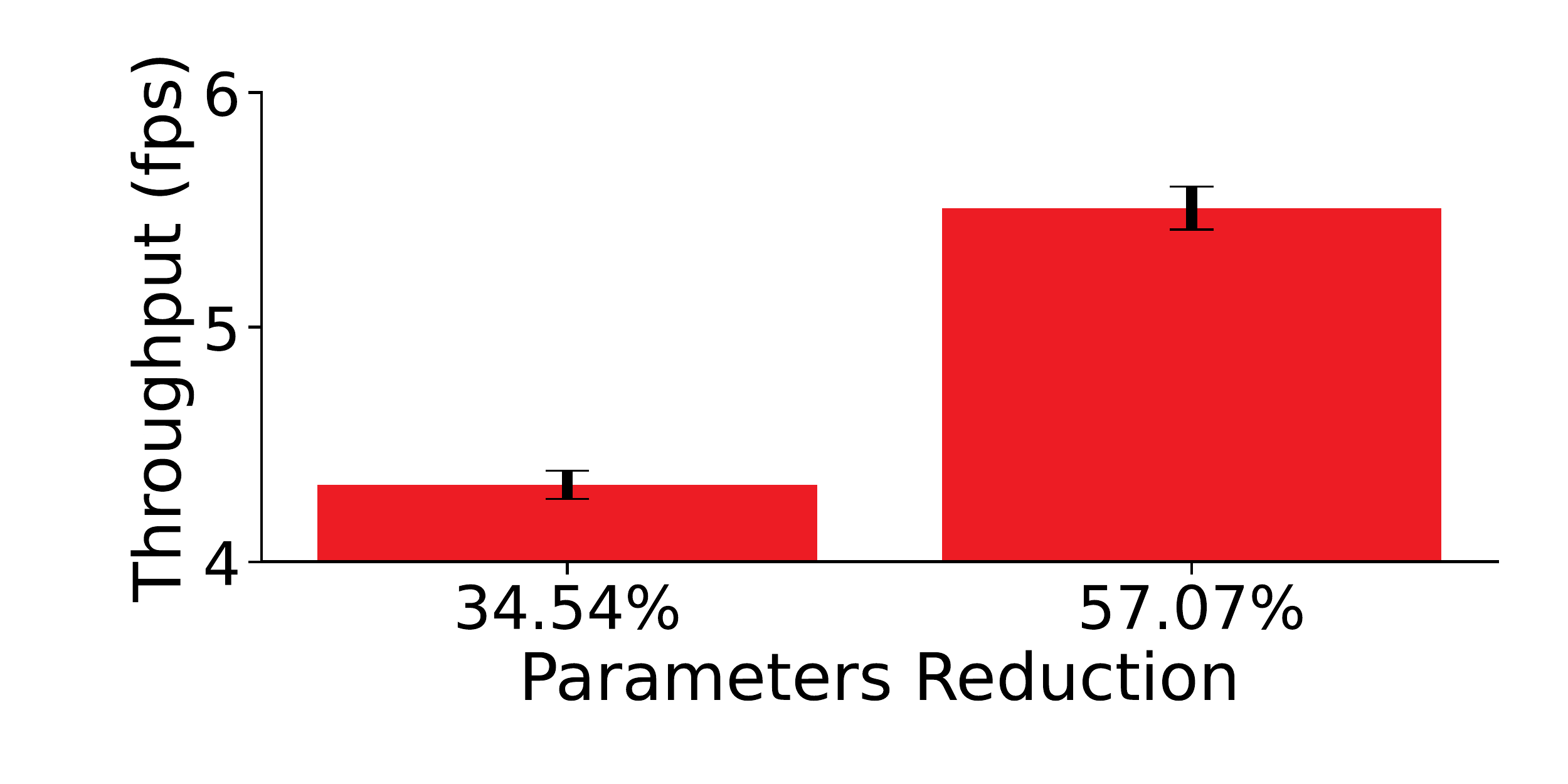}
        \caption{Throughput}
	\end{subfigure}
	\caption{The illustration of (a) the speedup and (b) the throughput of ResNet-50 with a parameter reduction of 34.54\% and 57.07\%, respectively, on one Intel(R) Xeon(R) Silver 4215R CPU. We run it for 10 trails. The input image size is $224 \times 224$. The average throughput of the original ResNet-50 is 3.67fps.}
 	\label{fig:speedup_cpu}
\end{figure*}

Figure~\ref{fig:speedup_cpu} shows the speedup of ResNet-50 with a parameter reduction of 34.54\% and 57.07\%, respectively, on one Intel(R) Xeon(R) Silver 4215R CPU. We run it for 10 trails. The input image size is $224 \times 224$. The average throughput of the original ResNet-50 for 10 trails is 3.67fps.


\subsubsection{Inference Speedup on Raspberry Pi}
\begin{figure*}[htp]
	\centering
	\begin{subfigure}{0.4\textwidth}
		\centering
  		\includegraphics[width=6.0cm]{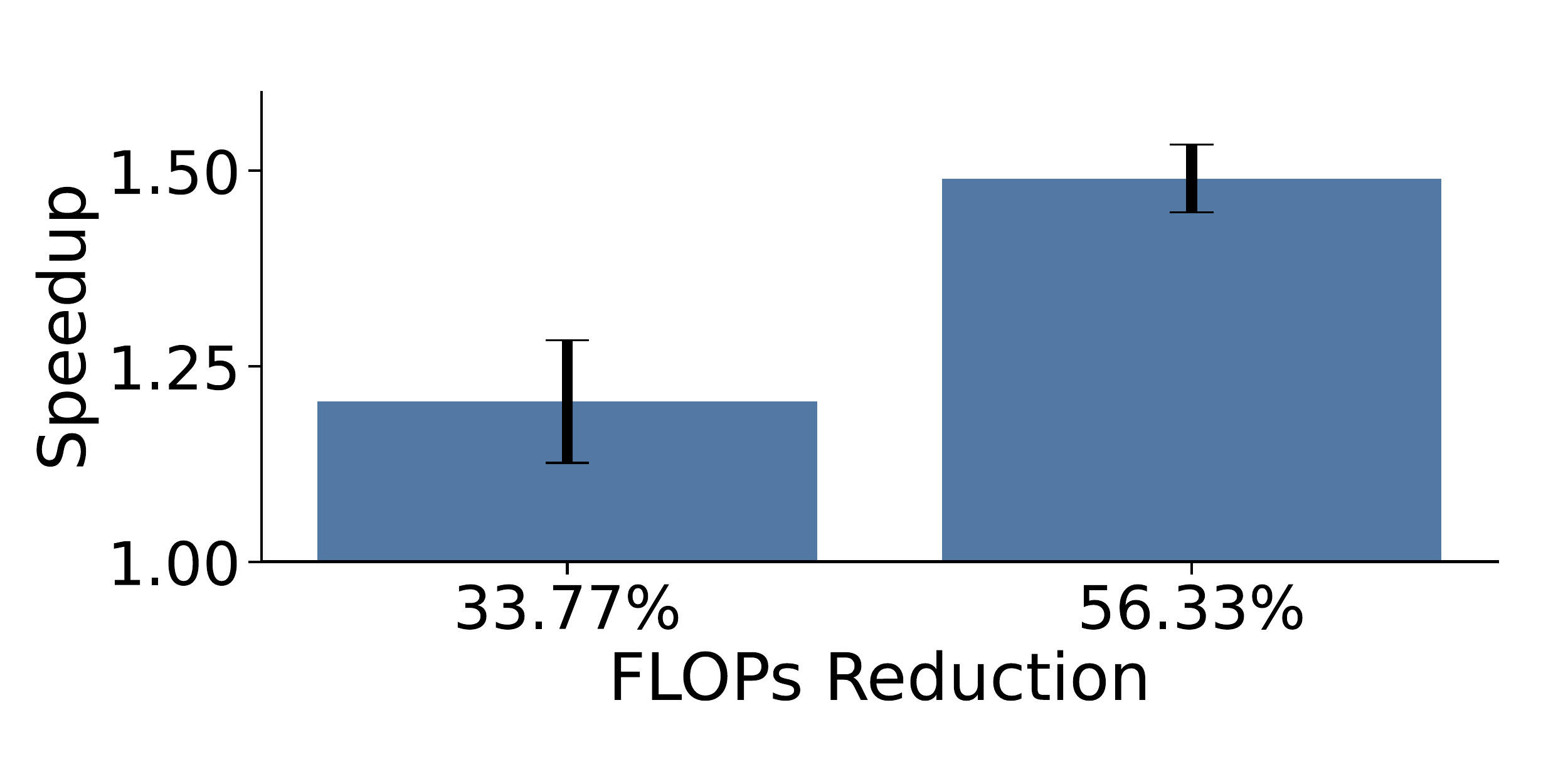}
		\caption{Speedup}
	\end{subfigure}
	\begin{subfigure}{0.4\textwidth}
		\centering
  		\includegraphics[width=6.0cm]{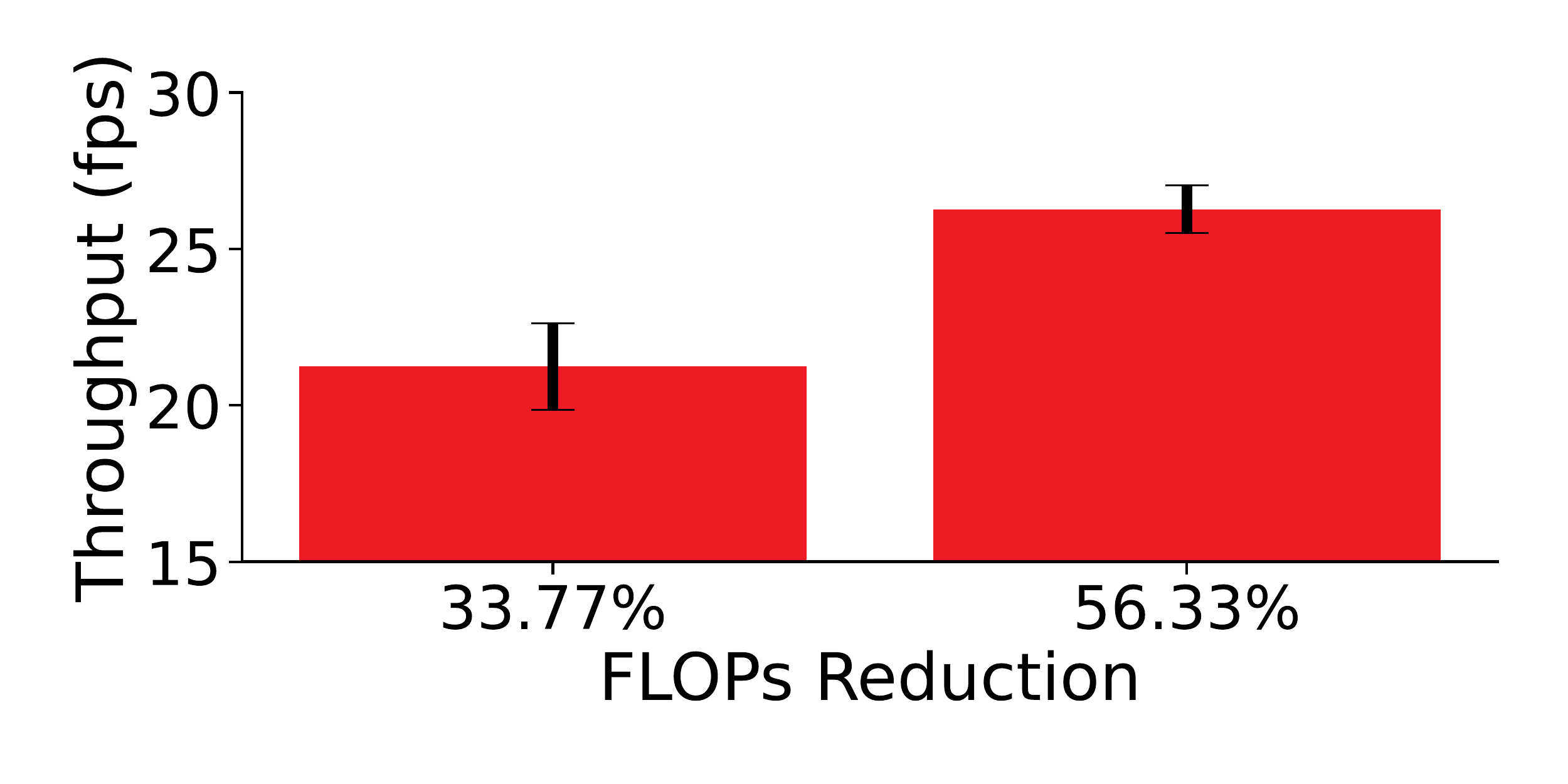}
		\caption{Throughput}
	\end{subfigure}
	\caption{The illustration of (a) the speedup and (b) the throughput of ResNet-56 with a FLOP reduction of 33.77\% and 56.33\%, respectively, on Raspberry Pi 3B+. We run it for 10 trails. The input image size is $32 \times 32$. The average throughput of the original ResNet-56 is 17.63fps.}
	\label{fig:speedup_pi_cifar10_rn56}
\end{figure*}

\boldhdr{ResNet-56 on CIFAR-10}
Figure~\ref{fig:speedup_pi_cifar10_rn56} shows the speedup of ResNet-56 with a FLOP reduction of 33.77\% and 56.33\%, respectively, on Raspberry Pi 3B+. We run it for 10 trails. The input image size is $32 \times 32$. The average throughput of the original ResNet-56 for 10 trails is 17.63fps.


\begin{figure*}[htp]
	\centering
	\begin{subfigure}{0.4\textwidth}
		\centering
  		\includegraphics[width=6.0cm]{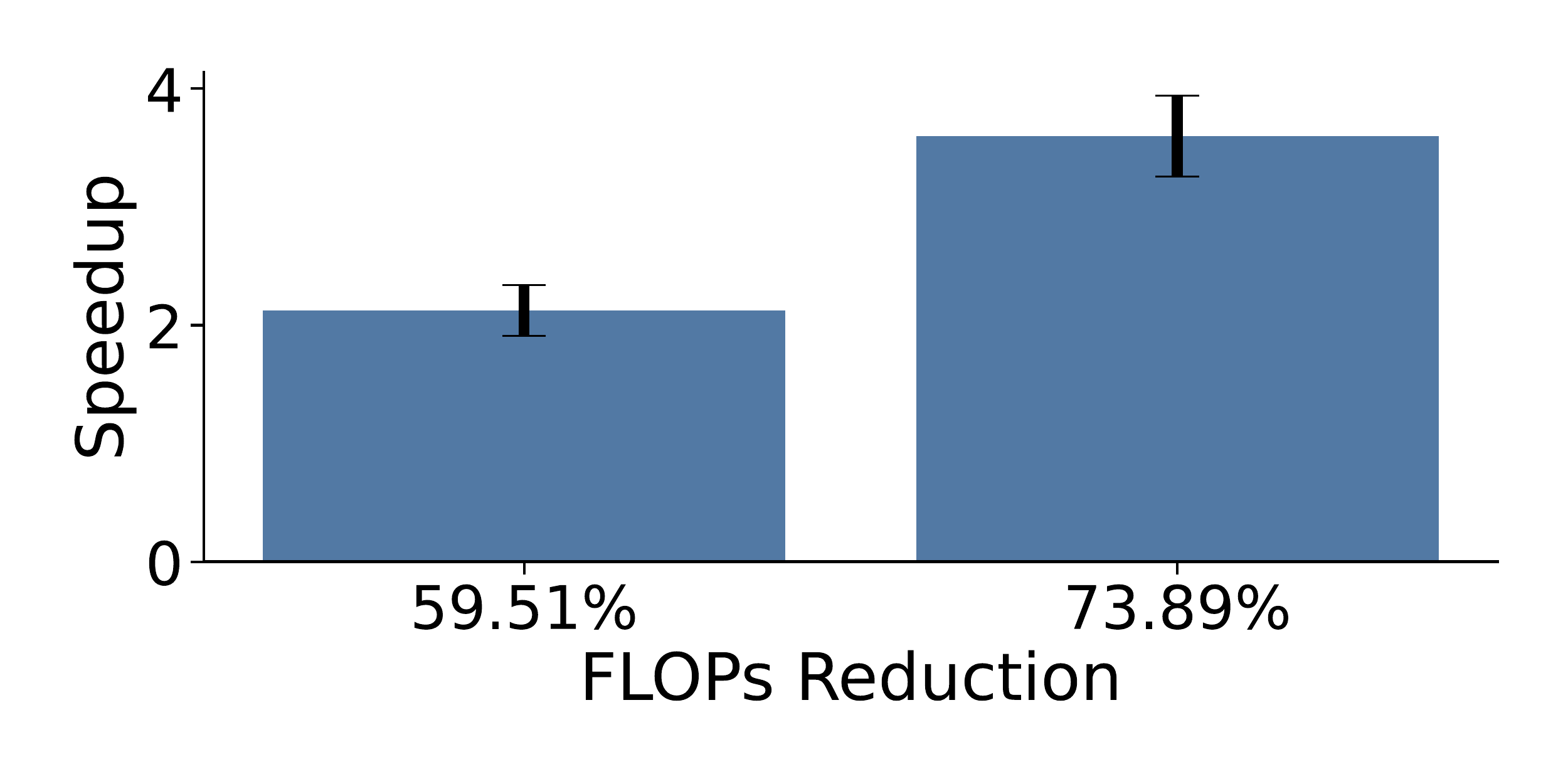}
		\caption{Speedup}
	\end{subfigure}
	\begin{subfigure}{0.4\textwidth}
		\centering
  		\includegraphics[width=6.0cm]{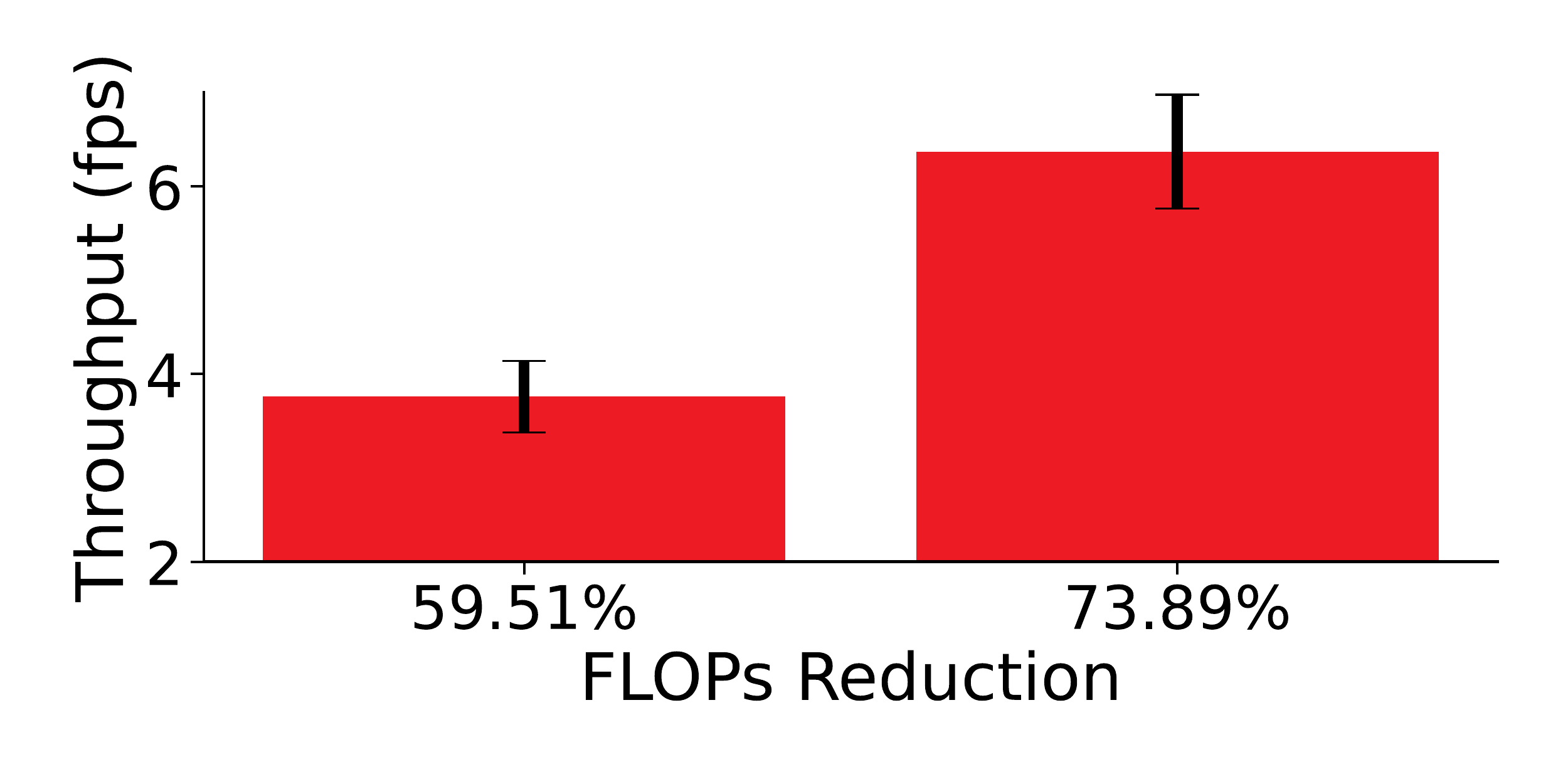}
		\caption{Throughput}
	\end{subfigure}
	\caption{The illustration of (a) the speedup and (b) the throughput of VGG-16 with a FLOP reduction of 59.51\% and 73.89\%, respectively, on Raspberry Pi 3B+. We run it for 10 trails. The input image size is $32 \times 32$. The average throughput of the original VGG-16 is 1.77fps.}
	\label{fig:speedup_pi_cifar10_vgg16}
\end{figure*}

\boldhdr{VGG-16 on CIFAR-10}
Figure~\ref{fig:speedup_pi_cifar10_vgg16} shows the speedup of VGG-16 with a FLOPs reduction of 59.51\% and 73.89\%, respectively, on Raspberry Pi 3B+. We run it for 10 trails. The input image size is $32 \times 32$. The average throughput of the original VGG-16 for 10 trails is 1.77fps.


\newpage
\subsubsection{Attention Distributions}
\begin{figure*}[h]
	\centering
	\includegraphics[width=\textwidth]{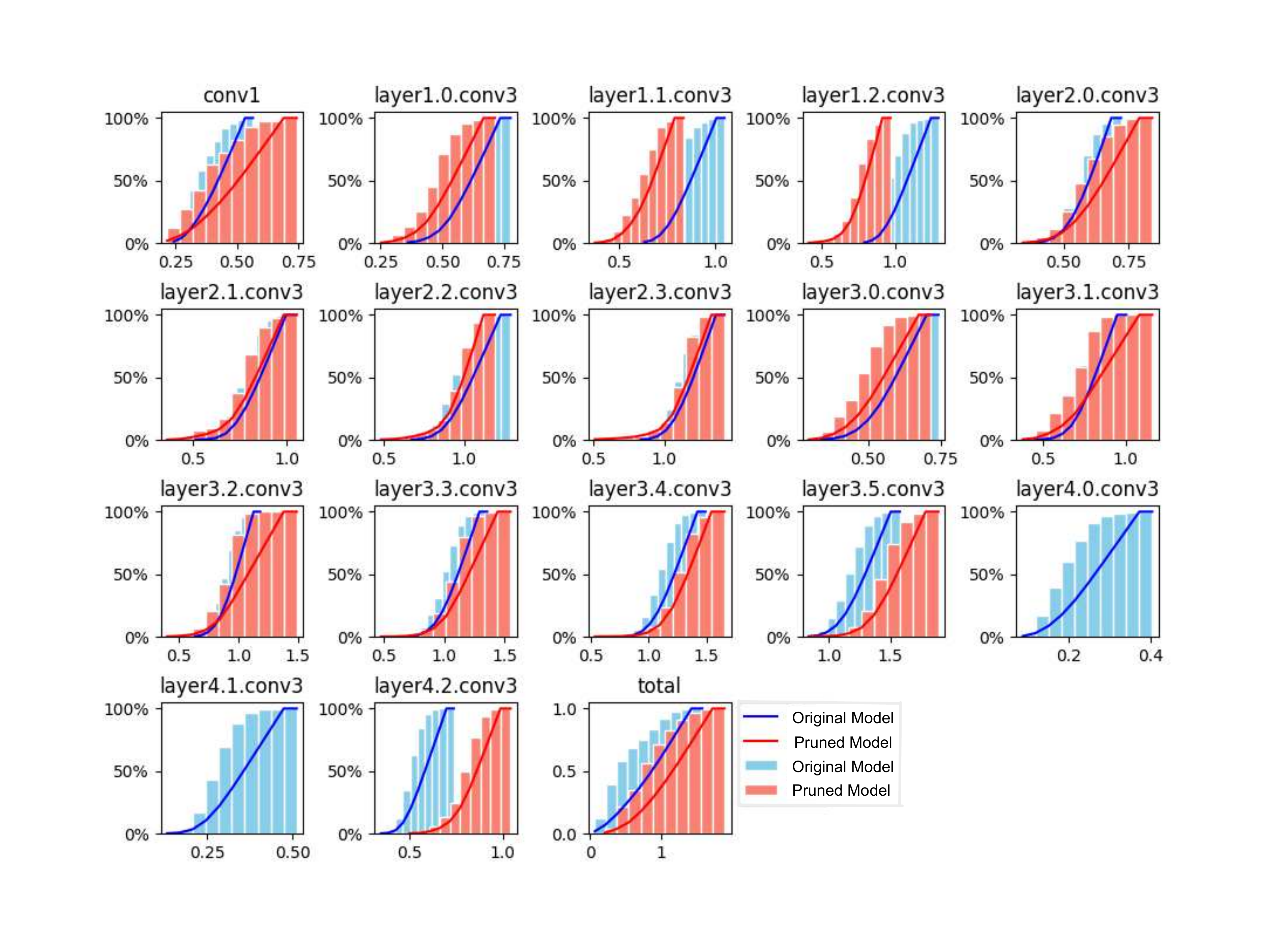}
	\caption{The the distribution of the attention values of each convolutional layer of the original ResNet-50 and the pruned ResNet-50 with a parameter reduction of 96.31\% on ImageNet.}
	\label{fig:distribution}
\end{figure*}

Figure~\ref{fig:distribution} shows the distribution of the attention values of each convolutional layer of the original ResNet-50 and pruned ResNet-50 with a parameter reduction of 96.31\% on ImageNet. Specifically, given one batch of images, the models do the inference once. We first measure the attention value ($p=1$) of each filter for each image, and then we calculate its average attention values for one batch of images.

\end{document}